\documentclass[journal]{IEEEtran}

\usepackage[utf8]{inputenc}
\usepackage{graphicx}
\usepackage{amssymb}
\usepackage{amsthm}
\usepackage{amsmath}
\usepackage{amsfonts}
\usepackage{tabularx,colortbl}
\usepackage{color}
\usepackage[english]{babel}
\usepackage{subfigure}
\usepackage{algorithm}
\usepackage{algorithmic}
\usepackage{balance}
\usepackage[compress]{cite}
\usepackage{mathrsfs}
\usepackage{todonotes}
\usepackage{multirow}
\usepackage{enumitem}
\usepackage{pstricks}
\usepackage{arydshln}
\newcommand{\mc}{\mathcal}
\newcommand{\R}{\mathbb{R}}
\newcommand{\norm}[2]{\left\Vert #1 \right\Vert_{{#2}}}
\DeclareMathOperator*{\argmin}{arg\,min}
\newcommand{\prob}{\mathbb{P}}
\newcommand{\expect}{\mathbb{E}}
\newcommand{\smean}{\overline}
\DeclareMathOperator{\var}{Var}

\newcommand{\frmu}{\mu}
\DeclareMathOperator{\frvar}{V_f}

\DeclareMathOperator{\image}{Im}
\newcommand{\dzvector}[1]{\mathbf{\boldsymbol{#1}}}
\renewcommand{\vec}[1]{\dzvector{#1}}
\renewcommand{\emptyset}{\varnothing}
\newcommand{\nomenclature}[2]{{#1} & {#2} \\}

\theoremstyle{plain}
\newtheorem{lemma}{Lemma}
\newtheorem{proposition}{Proposition}

\begin{document}

\title{Concept Drift and Anomaly Detection in Graph Streams}

\author{Daniele Zambon,~\IEEEmembership{Student Member,~IEEE},
        Cesare Alippi,~\IEEEmembership{Fellow,~IEEE},
        and Lorenzo Livi,~\IEEEmembership{Member,~IEEE}
\thanks{Manuscript received ; revised .}
\thanks{Daniele Zambon is with the Faculty of Informatics, 
Universit\`{a} della Svizzera italiana, 
Lugano, Switzerland 
(e-mail: daniele.zambon@usi.ch).}
\thanks{Cesare Alippi is with the Dept. of Electronics, Information, and Bioengineering, 
Politecnico di Milano, 
Milan, Italy and 
Faculty of Informatics, 
Universit\`{a} della Svizzera italiana, 
Lugano, Switzerland 
(e-mail: cesare.alippi@polimi.it, cesare.alippi@usi.ch)}
\thanks{Lorenzo Livi is with the Department of Computer Science, College of Engineering, Mathematics and Physical Sciences, 
University of Exeter, 
Exeter EX4 4QF, United Kingdom 
(e-mail: l.livi@exeter.ac.uk).}
}

\maketitle

\begin{abstract}
Graph representations offer powerful and intuitive ways to describe data in a multitude of application domains.
Here, we consider stochastic processes generating graphs  
and propose a methodology for detecting changes in stationarity of such processes.
The methodology is general and considers a process generating attributed graphs with a variable number of vertices/edges, without the need to assume one-to-one correspondence between vertices at different time steps.
The methodology acts by embedding every graph of the stream into a vector domain, where a conventional multivariate change detection procedure can be easily applied.
We ground the soundness of our proposal by proving several theoretical results.
In addition, we provide a specific implementation of the methodology and evaluate its effectiveness on several detection problems involving attributed graphs representing biological molecules and drawings.
Experimental results are contrasted with respect to suitable baseline methods, demonstrating the effectiveness of our approach.
\end{abstract}

\section*{Nomenclature}

\begin{center}\small
\begin{tabular}{lp{65mm}}
\nomenclature{$\mc G$}{Graph domain}
\nomenclature{$d(\cdot,\cdot)$}{Graph distance $\mc G\times \mc G\rightarrow \R_+$}
\nomenclature{$g_t$}{Generic graph in $\mc G$ generated at time $t$}
\nomenclature{$\vec g$}{Set $\{g_{t_1},\dots,g_{t_n}\}$ of graphs}
\nomenclature{$\mc D$}{Dissimilarity domain $\R^M$}
\nomenclature{$d'(\cdot,\cdot)$}{Distance $\mc D\times \mc D\rightarrow \R_+$}
\nomenclature{$y_t$}{Dissimilarity representation of generic graph $g_t$}
\nomenclature{$\vec y$}{Set $\{y_{t_1},\dots,y_{t_n}\}$ of diss. representations of $\vec g$}
\nomenclature{$\zeta(\cdot)$}{Dissimilarity representation $(\mc G,d)\rightarrow(\mc D,d')$}
\nomenclature{$R$}{Set of prototypes $\{r_1,\dots,r_M\}\subset\mc G$}
\nomenclature{$\mc P$}{Stochastic process generating graphs}
\nomenclature{$\tau, \widehat\tau$}{Change time and its estimate}
\nomenclature{$(\mc G,\mc S, Q)$}{Probability space for $\mc G$ in nominal regime}
\nomenclature{$(\mc D,\mc B, F)$}{Probability space for $\mc D$ in nominal regime}
\nomenclature{$H_0,H_1$}{Null and alternative hypotheses of a statistical test}
\nomenclature{$S_t$}{Statistic used by the change detection test}
\nomenclature{$h_t$}{Threshold for the change detection test}
\nomenclature{$\alpha_t$}{Significance level of the change detection test}
\nomenclature{$\smean{\vec y}, \expect[F]$}{Sample mean and expected value w.r.t.\ $F$}
\nomenclature{$\frmu[\vec g], \frmu[Q]$}{Fr\'echet sample and population means w.r.t.\ $Q$}
\end{tabular}
\end{center}

\section{Introduction}
\label{sec:intro}

Learning in non-stationary environments is becoming a hot research topic, as proven by the increasing body of literature on the subject, e.g., see \cite{7296710,gama2014survey} for a survey.
Within this learning framework, 
it is of particular relevance the detection of changes in stationarity of the data generating process. This can be achieved by means of either passive approaches \cite{elwell2011incremental}, which follow a pure on-line adaptation strategy, or active ones \cite{alippi2008justI,alippi2009just}, enabling learning only as a proactive reaction to a detected change in stationarity.
In this paper, we follow this last learning strategy, though many results are general and can be suitably integrated in passive learning approaches as well.

Most change detection mechanisms have been proposed for numeric i.i.d.\ sequences and either rely on change point methods or change detection tests.
Both change point methods and change detection tests are statistical tests; the former works off-line over a finite number of samples
\cite{hawkins2003changepoint} while the latter employs a sequential analysis of incoming observations \cite{basseville1993detection} to detect possible changes.
These techniques were originally designed for univariate normal distributed variables, and only later developments extended the methodology to non-Gaussian  distributions \cite{alippi2006adaptive,ross2012two} and multivariate datastreams \cite{zamba2006multivariate,golosnoy2009multivariate}.

A somehow related field to change detection tests is one-class classification (e.g., see \cite{eocc_mi} and references therein). There, the idea is to model only the nominal state of a given system and detect non-nominal conditions
(e.g., outliers, anomalies, or faults) by means of inference mechanisms.
However, one-class classifiers typically process data batches with no specific presentation order, while change detection problems are sequential in nature.

% A bit of motivation
The important role played, nowadays, by graphs as description of dynamic systems is boosting, also thanks to recent discoveries of theoretical frameworks for performing signal processing on graphs \cite{7407399,shuman2013emerging} and for analysing temporal (complex) networks \cite{holme2012temporal,holme2015modern,masuda2016guide}.
However, very few works address the problem of detecting changes in stationarity in streams (i.e., sequences) of graphs \cite{wilson2016modeling,barnett2016change,masoller2015quantifying}, and, to the best of our knowledge, none of them tackles the problem by considering a generic family of graphs (e.g., graphs with a varying number of vertices/edges, and arbitrary attributes on them).
In our opinion, the reason behind such a lack of research in this direction lies in the difficulty of defining a sound statistical framework for generic graphs that, once solved, would permit to also detect changes in time variance in time-dependent graphs.
In fact, statistics grounds on concepts like average and expectation, which are not standard concepts in graph domains.
Fortunately, recent studies \cite{jain2016geometry,jain2016statistical} have provided some basic mathematical tools that allow us to move forward in this direction, hence, addressing the problem of detecting changes in stationarity in sequences of graphs.

% Graph matching problem
A key problem in analysing generic graphs refers to assessing their dissimilarity, which is a well-known hard problem \cite{gm_survey}.
The literature proposes two main approaches for designing such a measure of dissimilarity \cite{foggia2012graph,emmert2016fifty,roy2014modeling}.
In the first case, graphs are analysed in their original domain $\mc G$, whereas the second approach consists of mapping (either explicitly or implicitly) graphs to numeric vectors.
A well-known family of algorithms used to assess dissimilarity between graphs relies on the Graph Edit Distance (GED) approach \cite{bunke1983inexact}.
More specifically, GED algorithms count and weight the edit operations that are needed in order to make two input graphs equal.
Differently, other techniques take advantage of kernel functions \cite{7947106}, spectral graph theory \cite{qiu2006graph,bai2015quantum}, or assess graph similarity by looking for recurring motifs \cite{costa2010fast}.
The computational complexity associated with the graph matching problem inspired researchers to develop heuristics and approximations, e.g., see \cite{fankhauser2011speeding,fischer2015approximation,bougleux2017graph} and references therein.

\subsection{Problem formulation}
\label{sec:prob-formulation}

In this paper, we consider sequences of \emph{attributed graphs}, i.e., directed or undirected labelled graphs characterized by a variable number of vertices and edges \cite{jain2016geometry}.
Attributed graphs associate vertices and edges with generic labels, e.g., scalars, vectors, categorical, and user-defined data structures.
In addition, multiple attributes can be associated to the same vertex/edge, whenever requested by the application.
By considering attributed graphs, we position ourselves on a very general framework covering most of application scenarios. However, generality requires a new operational framework, since all assumptions made in the literature to make the mathematics amenable, e.g., graphs with a fixed number of vertices and/or scalar attributes, cannot be accepted anymore. 
In order to cover all applications modellable through attributed graphs, we propose the following general problem formulation for change detection. 

Given a generic pre-metric distance $d(\cdot,\cdot)$ on $\mc G$, we construct a $\sigma$-algebra $\mc S$ containing at least all open balls of $(\mc G,d)$ and associate a generic probability measure $Q$ to $(\mc G ,\mc S)$.
The generated probability space $(\mc G ,\mc S, Q)$ allows us to consider graphs as a realisation of a structured random variable $g$ on $(\mc G ,\mc S, Q)$.
Define $\mc P$ to be the process generating a generic graph $g_t\in\mc G$ at time $t$ according to a stationary probability distribution $Q$ (nominal distribution).
We say that a change in stationarity occurs at (unknown) time $\tau$ when, from time $\tau$ on, $\mc P$ starts generating graphs according to a \emph{non-nominal} distribution $\widetilde Q\neq Q$, i.e.,
\begin{equation*}
 g_t \sim
\begin{cases}
 Q & t < \tau, \\
 \widetilde Q & t \geq \tau.
\end{cases}
\end{equation*}

In this paper, we focus on persistent (abrupt) changes in stationarity affecting the population mean.
However, our methodology is general and potentially can detect other types of change, including drifts and transient anomalies lasting for a reasonable lapse of time.

\subsection{Contribution and paper organization}

A schematic description of the proposed methodology to design change detection tests for attributed graphs is shown in Figure~\ref{fig:methodology}, and consists of two steps:
(i) mapping each graph $g_t$ to a numeric vector $y_t$ through a prototype-based embedding, and (ii) using a multivariate change detection test operating on the $y$-stream for detecting changes in stationarity.

The novelty content of the paper can be summarized as:
\begin{itemize}
\item A methodology to detect changes in stationarity in streams of \emph{attributed} graphs.
To the best of our knowledge, this is the first research contribution tackling change detection problems in streams of varying-size graphs with non-identified vertices and user-defined vertex/edge attributes;
\item A method derived from the methodology to detect changes in stationarity in attributed graphs. We stress that the user can design his/her own change detection method by taking advantage of the proposed methodology;
\item A set of theoretic results grounding the proposed methodology on a firm basis.
\end{itemize}

The proposed methodology is general and advances the few existing approaches for change detection in graph sequences mostly relying on the extraction and processing of topological features of fixed-size graphs, e.g., see \cite{ranshous2015anomaly}.  

It is worth emphasising that the proposed approach assumes neither one-to-one nor partial correspondence between vertices/edges across time steps (i.e., vertices do not need to be uniquely identified).
This fact has important practical implications in several applications.
As a very relevant example, we refer to the identification problem of neurons in extra-cellular brain recordings based on their activity \cite{Rossant2016}.
In fact, each electrode usually records the activity of a neuron cluster, and single neurons need to be disentangled by a procedure called spike sorting. Hence, a precise identification of neurons (vertices) is virtually impossible in such an experimental setting, stressing the importance of methods that do not require one-to-one correspondence between vertices over time.

The remainder of this paper is structured as follows.
Section~\ref{sec:related_work} contextualizes our contribution and discusses related works.
Section~\ref{sec:methodology} presents the proposed methodology for change detection in generic streams of graphs.
Theoretical results are sketched in Section \ref{sec:theoretical_results}; related proofs are given in Appendix \ref{sec:proofs-general}.
A specific implementation of the methodology is presented in Sections \ref{sec:our-implementation} and
related proofs in Appendix \ref{sec:proofs-particular}.
Section~\ref{sec:experiments} shows experimental results conducted on datasets of attributed graphs.
Finally, Section~\ref{sec:conclusions} draws conclusions and offers future directions.
Appendices \ref{app:F-is-prob-measure} and \ref{app:frechet-mean} provide further technical details regarding the problem formulation.

\section{Related work}
\label{sec:related_work}

The relatively new field of temporal networks deals with graph-like structures that undergo events across time \cite{holme2015modern,masuda2016guide}. 
Such events mostly realise in instantaneous or persistent \emph{contacts} between pairs of vertices. 
With such structures one can study dynamics taking place on a network, like epidemic and information spreading, and/or dynamics of the network itself, i.e., structural changes affecting vertices and edges over time.
Further relevant directions in temporal networks include understanding the (hidden) driving mechanisms and generative models \cite{holme2012temporal}.

The literature in statistical inference on time-varying graphs (or networks) is rather limited \cite{holme2015modern,jain2016statistical}, especially when dealing with attributed graphs and non-identified vertices.
Among the many, anomaly detection in graphs emerged as a problem of particular relevance, as a consequence of the ever growing possibility to monitor and collect data coming from natural and man-made systems of various size.
An overview of proposed approaches for anomaly and change detection on time-variant graphs is reported in \cite{ranshous2015anomaly,akoglu2015graph}, where the authors distinguish the level of influence of a change.
They identify changes affecting vertices and edges, or involving entire sub-networks of different size; this type of change usually concerns static networks, where the topology is often fixed.
Other changes have a global influence, or might not be ascribed to specific vertices or edges.

We report that there are several applications in which the vertices are labelled in such a way that, from a time step to another, we are always able to create a partial one-to-one correspondence (identified vertices). This case arises, e.g., when the identity of vertices plays a crucial role and must be preserved over time. 
Here, we put ourself in the more general scenario where vertices are not necessarily one-to-one identifiable through time.

Within the anomaly detection context, only few works tackle the problem in a classical change detection framework.
Among the already published works in detecting changes in stationarity, we mention Barnett et al. \cite{barnett2016change}, whose paper deals with the problem of monitoring correlation networks by means of a change point method. 
In particular, at every time step $t$, the authors construct the covariance matrix computed from the signals up to time $t$ and the covariance matrix of the remaining data.
As statistic for the change point model, they adopt the Frobenius norm between the covariance matrices. The authors evaluate their method on functional magnetic resonance imaging and stock returns.
A different way to approach the problem consists in modeling the network-generating process within a probabilistic framework. Graphs with $N$ vertices and disjoint communities can be described by the degree corrected stochastic block model, where some parameters represent the tendency of single vertices to be connected and communities to interact. This model has been adopted by Wilson et al.\ for monitoring the U.S.\ Senate co-voting network \cite{wilson2016modeling}. As monitoring strategy, they consider the standard deviation of each community, and then apply exponential weighted moving average control chart.
A further example of change point method for fixed-size graphs combines a generative hierarchical random graph model with a Bayesian hypothesis test \cite{peel2015detecting}.

\section{Change detection in a stream of graphs}

The structure of the section is as follows.
Section \ref{sec:methodology} describes the proposed methodology at a high level to ease the understanding.
In Section \ref{sec:theoretical_results}, we present theoretical results grounding our proposal; their proofs are given in the appendices. 
\begin{figure*}
\centering
\def\svgwidth{\textwidth}
\scriptsize
% \tiny

%% Creator: Inkscape inkscape 0.91, www.inkscape.org
%% Accompanies image file 'methodology.pdf' (pdf, eps, ps)
\begingroup%
  \makeatletter%
  \providecommand\color[2][]{%
    \errmessage{(Inkscape) Color is used for the text in Inkscape, but the package 'color.sty' is not loaded}%
    \renewcommand\color[2][]{}%
  }%
  \providecommand\transparent[1]{%
    \errmessage{(Inkscape) Transparency is used (non-zero) for the text in Inkscape, but the package 'transparent.sty' is not loaded}%
    \renewcommand\transparent[1]{}%
  }%
  \providecommand\rotatebox[2]{#2}%
  \ifx\svgwidth\undefined%
    \setlength{\unitlength}{595.27559055bp}%
    \ifx\svgscale\undefined%
      \relax%
    \else%
      \setlength{\unitlength}{\unitlength * \real{\svgscale}}%
    \fi%
  \else%
    \setlength{\unitlength}{\svgwidth}%
  \fi%
  \global\let\svgwidth\undefined%
  \global\let\svgscale\undefined%
  \makeatother%
  \begin{picture}(1,0.1952381)%
    \put(0,0){\includegraphics[width=\unitlength,page=1]{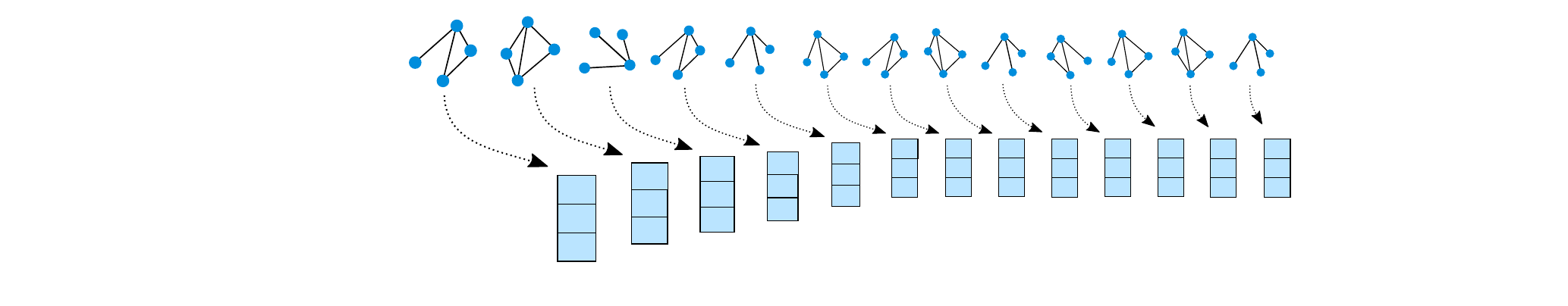}}%
    \put(0.28701589,0.12967157){\color[rgb]{0,0,0}\makebox(0,0)[lb]{\smash{$g_2$}}}%
    \put(0.34408731,0.13454381){\color[rgb]{0,0,0}\makebox(0,0)[lb]{\smash{$g_3$}}}%
    \put(0.35274579,0.08726144){\color[rgb]{0,0,0}\makebox(0,0)[lb]{\smash{$y_2$}}}%
    \put(0.39899008,0.0950862){\color[rgb]{0,0,0}\makebox(0,0)[lb]{\smash{$y_3$}}}%
    \put(0.30225668,0.10851184){\color[rgb]{0,0,0}\makebox(0,0)[lb]{\smash{$\zeta$}}}%
    \put(0.35548286,0.1141374){\color[rgb]{0,0,0}\makebox(0,0)[lb]{\smash{$\zeta$}}}%
%    \put(0.40209544,0.11758398){\color[rgb]{0,0,0}\makebox(0,0)[lb]{\smash{$\zeta$}}}%
%    \put(0.44954225,0.11968567){\color[rgb]{0,0,0}\makebox(0,0)[lb]{\smash{$\zeta$}}}%
%    \put(0.39009862,0.13772428){\color[rgb]{0,0,0}\makebox(0,0)[lb]{\smash{$g_4$}}}%
%    \put(0.44500139,0.09826667){\color[rgb]{0,0,0}\makebox(0,0)[lb]{\smash{$y_4$}}}%
    \put(0,0){\includegraphics[width=\unitlength,page=2]{methodology.pdf}}%
%    \put(0.44002931,0.13932907){\color[rgb]{0,0,0}\makebox(0,0)[lb]{\smash{$g_5$}}}%
%    \put(0.48921916,0.10136175){\color[rgb]{0,0,0}\makebox(0,0)[lb]{\smash{$y_5$}}}%
    \put(0.79868284,0.13681715){\color[rgb]{0,0,0}\makebox(0,0)[lb]{\smash{$g_t$}}}%
    \put(0.80849647,0.10981558){\color[rgb]{0,0,0}\makebox(0,0)[lb]{\smash{{\tiny$y_t$}}}}%
    \put(0.76101565,0.13669027){\color[rgb]{0,0,0}\makebox(0,0)[lb]{\smash{$g_{t-1}$}}}%
    \put(0.77438907,0.10979337){\color[rgb]{0,0,0}\makebox(0,0)[lb]{\smash{{\tiny$y_{t-1}$}}}}%
    \put(0.72185547,0.13682672){\color[rgb]{0,0,0}\makebox(0,0)[lb]{\smash{$g_{t-2}$}}}%
    \put(0.73808846,0.10987622){\color[rgb]{0,0,0}\makebox(0,0)[lb]{\smash{{\tiny$y_{t-2}$}}}}%
%    \put(0.6850596,0.13673895){\color[rgb]{0,0,0}\makebox(0,0)[lb]{\smash{$g_{t-3}$}}}%
%    \put(0.64378501,0.1370024){\color[rgb]{0,0,0}\makebox(0,0)[lb]{\smash{$g_{t-4}$}}}%
%    \put(0.60619875,0.13691463){\color[rgb]{0,0,0}\makebox(0,0)[lb]{\smash{$g_{t-5}$}}}%
%    \put(0.56949068,0.13695852){\color[rgb]{0,0,0}\makebox(0,0)[lb]{\smash{$g_{t-6}$}}}%
%    \put(0.59951847,0.10857622){\color[rgb]{0,0,0}\makebox(0,0)[lb]{\smash{$y_{t-6}$}}}%
%    \put(0.63314464,0.10863833){\color[rgb]{0,0,0}\makebox(0,0)[lb]{\smash{$y_{t-5}$}}}%
%    \put(0.66717378,0.10845201){\color[rgb]{0,0,0}\makebox(0,0)[lb]{\smash{$y_{t-5}$}}}%
%    \put(0.70405935,0.10851425){\color[rgb]{0,0,0}\makebox(0,0)[lb]{\smash{$y_{t-3}$}}}%
    \put(0,0){\includegraphics[width=\unitlength,page=3]{methodology.pdf}}%
    \put(0.25624518,0.09222956){\color[rgb]{0,0,0}\makebox(0,0)[lb]{\smash{$y_1$}}}%
    \put(0,0){\includegraphics[width=\unitlength,page=4]{methodology.pdf}}%
    \put(0.25932645,0.04049006){\color[rgb]{0,0,0}\makebox(0,0)[lb]{\smash{$d(g_1, r_2)$}}}%
    \put(0.25968972,0.06675576){\color[rgb]{0,0,0}\makebox(0,0)[lb]{\smash{$d(g_1, r_1)$}}}%
    \put(0.25922837,0.01441353){\color[rgb]{0,0,0}\makebox(0,0)[lb]{\smash{$d(g_1, r_3)$}}}%
    \put(0,0){\includegraphics[width=\unitlength,page=5]{methodology.pdf}}%
    \put(0.03308259,0.03891648){\color[rgb]{0,0,0}\makebox(0,0)[lb]{\smash{$R$:}}}%
    \put(0.03308259,0.02591648){\color[rgb]{0,0,0}\makebox(0,0)[lb]{\smash{set of}}}%
    \put(0.03308259,0.01591648){\color[rgb]{0,0,0}\makebox(0,0)[lb]{\smash{prototypes}}}%
%    \put(0.03308259,0.03891648){\color[rgb]{0,0,0}\makebox(0,0)[lb]{\smash{$R$: set of prototypes}}}%
    \put(0.194374,0.1094604){\color[rgb]{0,0,0}\makebox(0,0)[lb]{\smash{$\zeta:\mc G\rightarrow \mc D$}}}%
    \put(0.19239969,0.1326064){\color[rgb]{0,0,0}\makebox(0,0)[lb]{\smash{$g_1$}}}%
    \put(0.0203647,0.07295246){\color[rgb]{0,0,0}\makebox(0,0)[lb]{\smash{$\zeta$: dissimilarity representation}}}%
    \put(0,0){\includegraphics[width=\unitlength,page=6]{methodology.pdf}}%
    \put(0.79369507,0.04542756){\color[rgb]{0,0,0}\makebox(0,0)[lb]{\smash{change detection}}}%
    \put(0,0){\includegraphics[width=\unitlength,page=7]{methodology.pdf}}%
    \put(0.80603293,0.01495319){\color[rgb]{0,0,0}\makebox(0,0)[lb]{\smash{$S_w>h_w$ ?}}}%
    \put(0.90290668,0.11197745){\color[rgb]{0,0,0}\makebox(0,0)[lb]{\smash{no}}}%
    \put(0.90232817,0.08052135){\color[rgb]{0,0,0}\makebox(0,0)[lb]{\smash{yes}}}%
    \put(0.93134975,0.11334406){\color[rgb]{0,0,0}\makebox(0,0)[lb]{\smash{acquire}}}%
    \put(0.93134975,0.10334406){\color[rgb]{0,0,0}\makebox(0,0)[lb]{\smash{$g_{t+1}$}}}%
%    \put(0.93134975,0.11334406){\color[rgb]{0,0,0}\makebox(0,0)[lb]{\smash{aquire $g_{t+1}$}}}%
    \put(0.93166793,0.08513798){\color[rgb]{0,0,0}\makebox(0,0)[lb]{\smash{change}}}%
    \put(0.93166793,0.07513798){\color[rgb]{0,0,0}\makebox(0,0)[lb]{\smash{detected}}}%
%    \put(0.93166793,0.08513798){\color[rgb]{0,0,0}\makebox(0,0)[lb]{\smash{change detected}}}%
    \put(0,0){\includegraphics[width=\unitlength,page=8]{methodology.pdf}}%
    \put(0.73767353,0.03112204){\color[rgb]{0,0,0}\makebox(0,0)[lb]{\smash{$S_w$ }}}%
    \put(0,0){\includegraphics[width=\unitlength,page=9]{methodology.pdf}}%
    \put(0.6182751,0.03347207){\color[rgb]{0,0,0}\makebox(0,0)[lb]{\smash{time}}}%
    \put(0,0){\includegraphics[width=\unitlength,page=10]{methodology.pdf}}%
    \put(0.005,0.179){\color[rgb]{0,0,0}\makebox(0,0)[l]{\smash{$\mc P$:}}}%
    \put(0.005,0.162){\color[rgb]{0,0,0}\makebox(0,0)[l]{\smash{graph}}}%
    \put(0.005,0.151){\color[rgb]{0,0,0}\makebox(0,0)[l]{\smash{generating}}}%
    \put(0.005,0.140){\color[rgb]{0,0,0}\makebox(0,0)[l]{\smash{process}}}%
%    \put(0.13433525,0.18397714){\color[rgb]{0,0,0}\makebox(0,0)[rb]{\smash{$\mc P$: graph generating process}}}%
    \put(0,0){\includegraphics[width=\unitlength,page=11]{methodology.pdf}}%
    \put(0.11702896,0.02238119){\color[rgb]{0,0,0}\makebox(0,0)[lb]{\smash{$r_1$}}}%
    \put(0.14531407,0.01075858){\color[rgb]{0,0,0}\makebox(0,0)[lb]{\smash{$r_2$}}}%
    \put(0.16657858,0.02566305){\color[rgb]{0,0,0}\makebox(0,0)[lb]{\smash{$r_3$}}}%
  \end{picture}%
\endgroup%
\caption{
The diagram represents the fundamental steps of the methodology. 
At the top of the figure, the stochastic process $\mc P$ generates over time a stream of graphs $g_1,g_{2},g_{3},\dots$. 
The embedding procedure $\zeta(\cdot)$ is described in the bottom-left corner. The embedding of graph $g_t$ is computed by considering the dissimilarity $d(g_t,r_m)$ w.r.t.\ each prototype graph $r_m\in R$, and returns the embedded vector $y_t$ (here 3-dimensional) lying in the dissimilarity space $\mc D$. 
The embedding procedure proceeds over time and generates the multivariate vector stream $y_1,y_{2},y_{3},\dots$. 
A change detection method (bottom-right) is applied to the $w$-th window extracted from the $y$-stream to evaluate whether a change was detected or not, hence, iterating the procedure with the acquisition of a new graph.}
\label{fig:methodology}
\end{figure*}

\subsection{The proposed methodology}
\label{sec:methodology}

The methodology operates on an input sequence of attributed graphs $g_1,g_2,\dots,g_t,\dots\,\in \mc G$ and, as sketched in Figure~\ref{fig:methodology}, it performs two steps: (i) Map (embed) input graphs to a vector domain $\mc D=\R^M$.
Embedding is carried out by means of the dissimilarity representation $\zeta:\mc G\rightarrow\mc D$, which embeds a generic graph $g_t\in\mc G$ to a vector $y_t\in\mc D$.
(ii) Once a multivariate i.i.d.\ vector stream $y_1,y_2,\dots,y_t,\dots$ is formed, change detection is carried out by inspecting such a numerical sequence with the designer favourite method.
The two phases are detailed in the sequel.

\medskip
\subsubsection{Dissimilarity representation}

The embedding of a generic graph $g\in\mc G$ is achieved by computing the dissimilarity between $g$ and the prototype graphs in $R=\{r_1,\dots,r_M\}\subset \mc G$,
\begin{equation}
\label{eq:diss_embedding}
y=\zeta(g):=\left[d(g,r_1),\dots, d(g,r_M)\right]^{\top}.
\end{equation}
The vector $y$ is referred to as the dissimilarity representation of $g$. 
Set $R$ has to be suitably chosen to induce informative embedding vectors. 
For a detailed discussion about dissimilarity representations and possible ways to select prototypes, we suggest \cite{pkekalska+duin2005}.

In order to make the mathematics more amenable, here we assume $d(\cdot,\cdot)$ to be a metric distance; nevertheless, in practical applications, one can choose more general dissimilarity measures.

\medskip
\subsubsection{Multivariate vector stream}
\label{sec:methodology:vector}

At time step $t$ the process $\mc P$ generates graph $g_t$, and the map $\zeta(\cdot)$ embeds $g_t$ onto vector $y_t=\zeta(g_t)\in \mc D$,
inducing a multivariate stream $y_1,y_2,\dots,y_t,\dots$ whose elements lie in $\mc D$. Figure~\ref{fig:methodology} depicts the continuous embedding of graph process.

Under the nominal condition for process $\mc P$, graphs $\{g_t\}_{t<\tau}$ are i.i.d.\ and drawn from probability space $(\mc G,\mc S,Q)$. Consequently, also vectors $y_t\in\mc D$ are i.i.d..
We now define a second probability space $(\mc D,\mc B, F)$ associated with embedded vectors $y_t$; 
in particular, here we propose to consider for $\mc B$ the Borel's $\sigma$-algebra generated by all open sets in $\mc D$. $F$ is the push forward probability function of $Q$ by means of $\zeta(\cdot)$, namely
\begin{equation}\label{eq:F-from-Q}
F(B)= Q(\zeta^{-1}(B)), \qquad \forall\ B\in\mc B.
\end{equation}
With such a choice of $F$, we demonstrate in Appendix \ref{app:F-is-prob-measure} that $F$ is a probability measure on $(\mc D,\mc B)$.

\medskip
\subsubsection{Change detection test}
\label{sec:cd-test}

By observing the i.i.d.\ vector stream $y_1,y_2,\dots,y_t,\dots$ over time we propose a multivariate change detection procedure to infer whether a change has occurred in the vector stream and, in turn, in the graph stream.

The change detection test is the statistical hypothesis test
$$
\begin{array}{ll}
H_0:\quad \expect[S_t] = 0\\
H_1:\quad \expect[S_t] > 0
\end{array}
$$
where $0$ is the expected value during the nominal -- stationary -- regime and $\expect[\cdot]$ is the expectation operator. Statistic $S_t$, which is applied to windows of the vector stream, is user-defined and is requested to increase when the process $\mc P$ becomes non stationary.

Often, the test comes with a threshold $h_t$ so that if 
\begin{equation}\label{eq:discrimination-rule}
S_t>h_t\quad  \Rightarrow \quad \text{a change is detected}
\end{equation}
and the estimated change time is 
\begin{equation*}
\widehat \tau = \inf \{\,t\,:\, S_t>h_t\,\}.
\end{equation*}

Whenever the distribution of $S_t$ under hypothesis $H_0$ is available -- or can be estimated -- the threshold can be related to an
user-defined significance level $\alpha_t$ so that $\alpha_t=\prob(S_t>h_t|H_0)$.

\subsection{Theoretical results}
\label{sec:theoretical_results}

In this section, we show some theoretical results related to the methodology presented in Section~\ref{sec:methodology}. In particular, we prove the following claims:
\begin{enumerate}
\item[(C1)] Once a change is detected in the dissimilarity space according to a significance level $\alpha'$, then a change occurs in probability with significance level $\alpha$ also in the graph domain; 
\item[(C2)] If a change occurs in the graph domain having set a significance level $\alpha$, then, with a significance level $\alpha'$, a change occurs also in the dissimilarity space.
\end{enumerate}

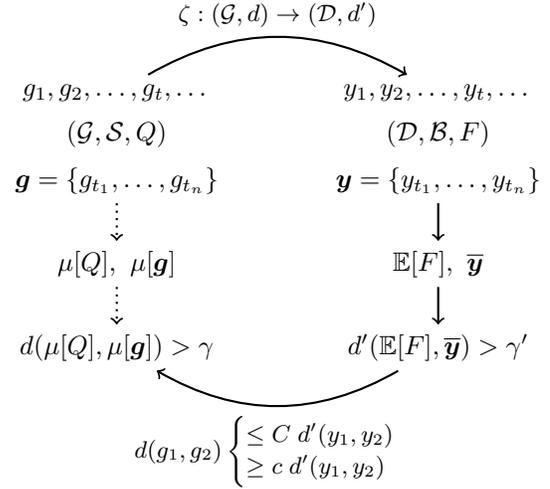
\begin{figure}
\centering
\newcommand{\wid}{1.5cm}
\newcommand{\hei}{.5cm}
\begin{tikzpicture}[->,auto,thick]
  \node (gs) [] {$g_1,g_2,\dots,g_t,\dots$};
  \node (ds) [right=\wid of gs] {$y_1,y_2,\dots,y_t,\dots$};
  \path[every node/.style={font=\sffamily\small}]
    (gs) edge[bend left] node [] {$\zeta:(\mc G,d)\rightarrow(\mc D,d')$} (ds);
  \node (gp) [below=0 of gs] {$(\mc G,\mc S,Q)$};
  \node (dp) [below=0 of ds] {$(\mc D ,\mc B,F)$};
  \node (gc) [below=0 of gp] {$\vec g=\{g_{t_1},\dots,g_{t_n}\}$};
  \node (dc) [below=0 of dp] {$\vec y=\{y_{t_1},\dots,y_{t_n}\}$};
  \node (g_) [below=\hei of gc] {$\frmu[Q],\ \frmu[\vec g]$};
  \node (d_) [below=\hei of dc] {$\expect[F],\ \smean{\vec y}$};
  \path[every node/.style={font=\sffamily\small}]
    (gc) edge[dotted] node {} (g_);
  \path[every node/.style={font=\sffamily\small}]
    (dc) edge node {} (d_);
  \node (gd) [below=\hei of g_] {$d(\frmu[Q],\frmu[\vec g])> \gamma$};
  \node (dd) [below=\hei of d_] {$d'(\expect[F],\smean{\vec y})> \gamma'$};
  \path[every node/.style={font=\sffamily\small}]
    (g_) edge[dotted] node {} (gd);
  \path[every node/.style={font=\sffamily\small}]
    (d_) edge node {} (dd);
  \path[every node/.style={font=\sffamily\small}]
    (dd) edge[bend left] node [] {$d(g_1,g_2)\begin{cases}\leq C\;d'(y_1,y_2)\\\geq c\;d'(y_1,y_2)\end{cases}$} (gd);
\end{tikzpicture}
\caption{Conceptual scheme of the proposed methodology that highlights the theoretical results. 
The key point is to show that objects close in the graph domain $\mc G$, onto which metric $d(\cdot,\cdot)$ is defined, under certain conditions, are close also in the embedded domain $\mc D$ controlled by metric $d'(\cdot,\cdot)$.
It comes up that if objects in the embedding domain $\mc D$ are distant in probability, then also the related graphs in $\mc G$ are distant, hence indicating a change in stationarity according to a given false positive rate.}
\label{fig:scheme}
\end{figure}

Figure~\ref{fig:scheme} depicts the central idea behind the methodology.
Through transformation $\zeta(\cdot)$, we map graphs to vectors.
In the transformed space, we consider the expectation $\expect[F]$ and the sample mean $\smean{\vec y}$ associated with set $\vec y=\{y_{t_1},\dots,y_{t_n}\}$ obtained by embedding the graph set $\vec g=\{g_{t_1},\dots,g_{t_n}\}$, 
and design a hypothesis test of the form 
\begin{equation}\label{eq:hp-test-d'}
\begin{array}{ll}
H_0:&d'(\expect[F],\smean{\vec y})\leq \delta\\
H_1:&d'(\expect[F],\smean{\vec y})>\delta
\end{array}
\end{equation}
where $\delta$ is a positive threshold and hypothesis $H_0$ is associated with a nominal, change-free condition.
In this paper, we relate the $y$-test of \eqref{eq:hp-test-d'} to a correspondent test $d(\frmu[Q],\frmu[\vec g])>\gamma$ operating in the graph domain, where $\frmu[Q],\frmu[\vec g]$ are, respectively, the population and sample mean defined according to Fr\'echet \cite{frechet1948elements}; for further details about Fr\'echet statistics refer to Appendix~\ref{app:frechet-mean}.

Define $\alpha$ and $\alpha'$ to be two significance levels, such that:
\begin{equation}
\label{eq:gammas-alphas}
\begin{array}{ll}
\alpha &=\prob(d(\frmu[Q],\frmu[\vec g])>\gamma|H_0),\\
\alpha'&=\prob(d'(\expect[F],\smean{\vec y})>\gamma'|H_0).
\end{array}
\end{equation}
In the sequel we relate the threshold $\gamma$ to $\gamma'$, so that also the significance levels $\alpha$ and $\alpha'$ are in turn related to each other.

In order to address our problem, we introduce mild assumptions to obtain closed-form expressions.
Such assumptions are satisfied in most applications. More specifically,
\begin{itemize}
\item[(A1)] We assume that the attributed graph space $(\mc G, d)$ and the dissimilarity space $(\mc D, d')$ are metric spaces; in particular, $(\mc G,d)$ is chosen as a graph alignment space \cite{jain2016statistical} -- i.e., a general metric space of attributed graphs -- and $d'(\cdot,\cdot)$ has to be induced by a norm.
\item[(A2)] We put ourselves in the conditions of \cite{jain2016statistical} in order to take advantage of results therein; specifically, we assume that the Fr\'echet function $\mc F_Q(g)=\int_{\mc G}d^2(g,f)\,dQ(f)$ is finite for any $g\in\mc G$, and there exists a sufficiently asymmetric graph $f$ such that the support of $Q$ is contained in a cone around $f$. In this way, we are under the hypotheses of Theorems 4.1 and 4.2 of \cite{jain2016statistical}, which ensure the existence and uniqueness of the Fr\'echet population and sample mean in $\mc G$.
\item[(A3)] The embedding function $\zeta:(\mc G,d)\rightarrow(\mc D,d')$ is bilipschitz, i.e., there exist two constants $c,C>0$, such that for any pair $g,f\in\mc G$
\begin{align}
\label{eq:lb}d(g,f) &\geq c\,d'(\zeta(g),\zeta(f))\\
\label{eq:ub}d(g,f) &\leq C\,d'(\zeta(g),\zeta(f)).
\end{align}
\end{itemize}

Let $\Psi(\cdot)$ and $\Upsilon(\cdot)$ be the cumulative density functions (CDFs) of $d(\frmu[\vec g], \frmu[Q])$ and $d'(\smean{\vec y}, \expect[F])$, respectively.
Proposition~\ref{prop:independent-bounds} bounds the distribution $\Psi(\cdot)$ in terms of $\Upsilon(\cdot)$. This fact yields the possibility to derive significance levels $\alpha,\alpha'$ and thresholds $\gamma,\gamma'$ in Equation \eqref{eq:gammas-alphas} that are related. 
%, i.e., $\alpha=1-\Psi(\gamma)$ and $\alpha'=1-\Psi(\gamma')$. 

In order to prove Proposition~\ref{prop:independent-bounds}, we make use of two auxiliary results, Lemmas~\ref{lemma:mean-and-zetamean} and \ref{lemma:meanLS-Phi-ell-u}.
At first we need to comment that, although in general $\smean{\vec y} \neq \zeta(\frmu[\vec g])$ and $\expect[F] \neq \zeta(\frmu[Q])$, differences are bounded in practice, as shown by Lemma~\ref{lemma:mean-and-zetamean}.
\begin{lemma}\label{lemma:mean-and-zetamean}
Considering a set $\vec g$ of i.i.d.\ random graphs and the associated embedded set $\vec y$, there exists a constant $v_2$ such that
$$\begin{array}{cl}
\norm{\expect[F] - \zeta\left(\frmu[Q]\right)}{2}^2       
\ &\leq\ v_2\\
\prob\left(\norm{\smean{\vec y} - \zeta(\frmu[\vec g])}{2}^2 \geq \delta \right)\ &\leq\ \frac{v_2}{\delta} \qquad \forall \delta>0.
\end{array}$$
\end{lemma}
Lemma~\ref{lemma:meanLS-Phi-ell-u} is used to derive bounds on the marginal distributions from bounds on the joint distributions 
%-- like \eqref{eq:lb} and \eqref{eq:ub} in Assumption (A3) -- 
that are useful in Proposition~\ref{prop:independent-bounds} to relate the threshold and the significance level in the graph space with the ones in the dissimilarity space.
\begin{lemma}\label{lemma:meanLS-Phi-ell-u}
Consider a random variable $x\in\mc X$ and two statistics $d_1(\cdot),d_2(\cdot):\mc X\rightarrow \R_+$ with associated CDFs $\Phi_1(\cdot)$, $\Phi_2(\cdot)$, respectively. If function $u:\R_+\rightarrow\R_+$ is increasing and bijective and $p$ is a constant in $[0,1]$, then %$\forall \gamma\geq0$
\begin{multline*}
\prob\left(d_1(x)\leq u(d_2(x))\right)\geq p
\ \Rightarrow\ 
\Phi_1(\cdot) \geq p\cdot\Phi_2(u^{-1}(\cdot)).
\end{multline*}
\end{lemma}

Claims (C1) and (C2) now follow from the relation between the CDFs $\Psi(\cdot)$ and $\Upsilon(\cdot)$ proven by the subsequent Proposition~\ref{prop:independent-bounds}.
In particular, regarding Claim (C1) Proposition~\ref{prop:independent-bounds} provides a criterion for setting a specific threshold $\gamma'$ for the test \eqref{eq:hp-test-d'} for which we can state that $d(\frmu[Q],\frmu[\vec g])$ is unexpectedly large with a significance level at most $\alpha$; similarly, we obtain Claim (C2).
\begin{proposition}\label{prop:independent-bounds}
Consider a sample $\vec g$ of i.i.d.\ graphs.
Under assumptions (A1)--(A3), if $\Psi(\cdot)$ and $\Upsilon(\cdot)$ are the CDFs of statistics $d(\frmu[Q],\frmu[\vec g])$ and $d'(\expect[F], \smean{\vec y})$, respectively, then for every $\delta>0$ there exist two values $b_\delta$ and $ p_\delta$, depending on $\delta$ but independent of $\vec g$, such that
% \begin{multline*}
$$
{ p_\delta} \cdot\Upsilon\left(\tfrac{\gamma}{C}-b_\delta\right)
\quad \leq\quad 
\Psi(\gamma) 
\quad \leq\quad 
\tfrac{1}{ p_\delta} \cdot\Upsilon\left(\tfrac{\gamma}{c}+b_\delta\right).
% \end{multline*}
$$
\end{proposition}
The proofs are given in Appendix~\ref{sec:proofs-general}. 
Results of Proposition~\ref{prop:independent-bounds} allows us to state the major claim (C1): given a sample $\vec g$, 
for any significance level $\alpha$ and threshold $\gamma$, as in \eqref{eq:gammas-alphas}, 
we can set up a $\gamma'$ such that
the confidence level of detecting a change in $\mc D$ is at least $ p_\delta(1-\alpha)$.
Specifically, for $\gamma'=\frac{\gamma}{c}+b_\delta$ we have that 
\begin{equation}\label{eq:claim:ci}
\prob\left(\,d'(\expect[F],\smean{\vec y})\leq\gamma'\,|\,H_0\,\right)=\Upsilon\left(\tfrac{\gamma}{c}+b_\delta\right)\geq  p_\delta(1-\alpha). %\\
\end{equation}
Equation \eqref{eq:claim:ci} states that if no
change occurs in $\mc G$ with confidence level $1-\alpha$, then a change will not be detected in $\mc D$ according to threshold $\gamma'$ at least with confidence level $ p_\delta(1-\alpha)$.
Indeed, this proves (C1) by contraposition.
Similarly, Proposition~\ref{prop:independent-bounds} allows to prove Claim (C2). In fact, for any $\alpha'$ and $\gamma'$ as in \eqref{eq:gammas-alphas}, we can set $\gamma$ so that $\gamma'=\frac{\gamma}{C}-b_\delta$ and obtain
\begin{align*}
\prob\left(\,d(\frmu[Q],\frmu[\vec g])\leq\gamma\,|\,H_0\,\right)=\Psi(\gamma)
\geq  p_\delta(1-\alpha').
\end{align*}
% where $\gamma$ is such that $\Upsilon\left( \frac{\gamma}{C}-b_\delta \right)\geq 1-\alpha'$.

\section{Implementations of the methodology}
\label{sec:our-implementation}

This section describes two examples showing how to implement the proposed methodology and derive actual change detection tests.
In Section \ref{sec:clt-implementation} we present a specific method for generic families of graphs, whereas in Section \ref{sec:identified-vertices-01} we further specialize the methodological results to a special case considering graphs with identified vertices.

\subsection{Change detection based on central limit theorem}
\label{sec:clt-implementation}

Here, we consider specific techniques for prototype selection and change detection test.
For the sake of clarity, we keep the subsection structure of Section~\ref{sec:methodology}. 
Both the prototype selection and change detection test require a training phase. For this reason, the first observed graphs of the stream will serve as training set $T$, that we assume to be all drawn under nominal conditions. 

\medskip
\subsubsection{Dissimilarity representation}

Since the change detection method operates in the dissimilarity space, we need to define the embedding $\zeta(\cdot)$ that, at each time step, maps a generic graph $g_t$ to a vector $y_t$.

We comment that the embedding $\zeta(\cdot)$ is completely defined once the graph distance metric $d(\cdot,\cdot)$ and prototype set $R$ are chosen.
Here, we adopt a metric GED as graph distance, 
{since it meets Assumption (A1) of $(\mc G,d)$ being a graph alignment space.} 

Many approaches have been proposed to deal with the prototype selection problem, e.g., see \cite{riesen+bunke2010} and references therein.
While the proposed methodology is general and one can choose any solution to this problem, here we adopt the k-Centres method \cite{riesen2007graph}.
The method selects prototypes so as to cover training data with balls of equal radius.
Specifically, the algorithm operates as follows:
\begin{enumerate} 
\item select $M$ random prototypes $R:=\{r_1,\dots,r_M\}\subseteq T$;
\item \label{algo:start-iter} for each $r_m\in R$, consider the set $C_m$ of all $g\in T$ such that $d(r_m,g)=\min_{r\in R} d(r,g)$;
\item for $m=1,\dots,M$ update the prototypes $r_m$ with a graph ${c\in T}$ minimising $\max_{g\in C_m} d(c,g)$;  
\item if the prototype set $R$ did not change from the previous iteration step then exit, otherwise go to step \ref{algo:start-iter}.
\end{enumerate}

In order to improve the robustness of the k-Centres algorithm, we repeat steps 1--4 by randomizing initial conditions and select the final prototype set $R$ to be 
$$
R=\argmin_{R\in \{R^{(i)}\}}\left\{\max_{r_m\in R} \left[ \max_{c\in C_m} d(r_m,c) \right]\right\},
$$
where $\{R^{(i)}\}$ is the collection of prototype sets found at each repetition.

\medskip
\subsubsection{Multivariate vector stream}

Every time the process $\mc P$ generates a graph $g_t$, we embed it as $y_t=\zeta(g_t)$ by using the prototype set $R$ identified with the k-Centres approach. 
This operation results in a multivariate vector stream $y_1,y_2,\dots,y_t,\dots$ on which we apply the change detection test.

\medskip
\subsubsection{Change detection test}

We consider here a variation of the cumulative sum (CUSUM) test \cite{alippi2006adaptive} to design the change detection test. CUSUM is based on the cumulative sum chart \cite{page1954continuous}, it has been proven to be asymptotically optimal \cite{basseville1993detection} and allows for a simple graphical interpretation of results \cite{manly2000cumulative}. 
Here, the CUSUM is adapted to the multivariate case. However, we remind that, in principle, any change detection test can be used on the embedded sequence. 

We batch the observed embedding vectors into non-overlapping samples $\vec y_w := \{y_{(w-1)n+1},\dots,y_{wn}\}$ of length $n$, where index $w$ represents the $w$-th data window.
For each window, we compare the sample mean $\smean{\vec y}_0$ estimated in the training set $T$ with that estimated in the $w$-th window, i.e., $\smean{\vec y}_w$ and compute the discrepancy
$$
s_w:=d'(\smean{\vec y}_0,\smean{\vec y}_w).
$$

By assuming that $y_1, y_2,\dots,y_t,\dots$ are i.i.d., and given sufficiently large $|T|$ and $n$, the central limit theorem grants that $\smean{\vec y}_0$ and $\smean{\vec y}_w$ are normally distributed.
In particular, $\smean{\vec y}_0$ and $\smean{\vec y}_w$ share the same expectation $\expect[F]$, while covariance matrices are $\frac{1}{|T|}\var[F]$ and $\frac{1}{{n}}\var[F]$, respectively.

As a specific choice of $d'(\cdot,\cdot)$, we adopt the Mahalanobis' distance, i.e., $d'(\smean{\vec y}_0,\smean{\vec y}_w):=d_\Sigma(\smean{\vec y}_0,\smean{\vec y}_w)$ where 
\begin{equation}
\label{eq:mahal-y0-yw}
d_\Sigma(\smean{\vec y}_0,\smean{\vec y}_w)=\sqrt{(\smean{\vec y}_0-\smean{\vec y}_w)^\top\Sigma^{-1}(\smean{\vec y}_0-\smean{\vec y}_w)},
\end{equation}
with matrix $\Sigma=\left(\frac{1}{|T|}+\frac{1}{n} \right)\var[F]$, i.e., the covariance matrix of $\smean{\vec y}_0-\smean{\vec y}_w$.
In our implementation, we consider as covariance matrix $\var[F]$ the unbiased estimator $\frac{1}{|T|-1}\sum_{y\in T} (y-\smean{\vec y}_0)\cdot(y-\smean{\vec y}_0)^\top$.

For each stationary window $w$, the squared Mahalanobis' distance $s_w^2$ is distributed as a $ \chi^2_M$.

The final statistic $S_w$ inspired by the CUSUM test is defined as 
\begin{equation}\label{eq:cusum-iteration}
\begin{cases}
S_w = \max_{}\left\{\;0\,,\;S_{w-1}+(s_w-q)\;\right\}\\
S_0 = 0.
\end{cases}
\end{equation}

The difference $s_w-q$ increases with the increase of the discrepancy in \eqref{eq:mahal-y0-yw} and positive values support the hypothesis that a change occurred, whereas negative values suggest that the system is still operating under nominal conditions. This behaviour resembles that of the original CUSUM increment associated with the log-likelihood ratio.
In particular, the parameter $q$ can be used to tune the sensitivity of the change detection; in fact, if $q^2$ is the $\beta$-quantile $\chi^2_M(\beta)$, then $s_w-q$ produces a negative increment with probability $\beta$ and a positive one with probability $1-\beta$.

The last parameter to be defined is the threshold $h_w$ as requested in \eqref{eq:discrimination-rule}. Since the non-nominal distribution is unknown, a common criterion suggests to control the rate of false alarms, $\alpha$.
In sequential detection tests, a related criterion requires a specific \emph{average run length} under the null hypothesis (ARL0) \cite{frisen2003statistical}.
ARL0 is connected to the false positive rate in the sense that setting a false alarm rate to $\alpha$ yields an ARL0 of $\alpha^{-1}$.
Since we propose a sequential test, statistic $S_w$ depends on statistics at preceding times.
As a consequence, since we wish to keep $\alpha$ fixed, we end up with a time-dependent threshold $h_w$. 
As done in \cite{zamba2006multivariate,golosnoy2009multivariate}, we numerically determine thresholds through Monte Carlo sampling by simulating a large number of processes of repeated independent $\chi^2_M$ realisations.
Threshold $h_w$ is then the $1-\alpha$ quantile of the estimated distribution of $S_w$.

We point out that, when setting a significance level for the random variable $S_w$, we are implicitly conditioning to the event $S_i\leq h_i,\ \forall i<w$; in fact, when $S_w$ exceeds $h_w$, we raise an alarm and reconfigure the detection procedure.

\medskip
\subsubsection{Theoretical results}

The choice of Mahalanobis' distance ensures that almost all assumptions in Section~\ref{sec:theoretical_results} are met.
In particular, the Mahalanobis' distance meets the requirements of Assumption (A1). 
Then, the following Lemma~\ref{lemma:d-mahalanobis-d-graph} provides a lower bound of the form \eqref{eq:lb} in Assumption (A3); specifically, the lemma shows that, up to a positive factor, the distance between two graphs is larger then the one between the associated embedding vectors.

From these properties, we can apply Proposition~\ref{prop:independent-bounds} to state that Claim (C1) holds. Hence for any $\alpha$, we can set a specific threshold $\gamma'$ yielding a confidence level at least $p_\delta(1-\alpha)$.
\begin{lemma}\label{lemma:d-mahalanobis-d-graph}
For any two graphs $g,f\in\mc G$, we have that
$$
d(g,f) 
\ \geq\  
\sqrt{\frac{\lambda_M}{M}}\; d_{\Sigma}(\zeta(g),\zeta(f)),
$$
where $\lambda_M$ is the smallest eigenvalue of $\Sigma$.
\end{lemma}
The proof is reported in Appendix~\ref{sec:proofs-particular}.
Distance $d_\Sigma(\cdot,\cdot)$ is well-defined only when $\Sigma$ is positive definite, a condition implying that selected prototypes are not redundant.

\subsection{A special case: graphs with identified vertices}
\label{sec:identified-vertices-01}

Here we take into account the particular scenario where the attribute function $a(\cdot)$ of $g=(V,E,a)\in\mc G$ assigns numerical attributes in $[0,1]$ to vertices and edges of $g$; the vertex set $V$ is a subset of a predefined finite set $\mc V$, with $|\mc V|=N$.
The peculiarity of this space $\mc G$ resides in the fact that any vertex permutation of a graph leads to a different graph.
Many real-world applications fall in this setting, for instance correlation graphs obtained by signals coming from sensors or graphs generated by transportation networks.

We show an example of method for this setting, which complies with the methodology and satisfies Assumption (A3).
This fact follows from the existence of an injective map $\omega$ from $\mc G$ to the $[0,1]^{N\times N}$ matrix set.
Indeed, we represent each graph with its weighted adjacency matrix whose row/column indices univocally correspond to the vertices in $\mc V$.
By endowing $\mc G$ with the Frobenius distance\footnote{In $\mc G$, $d_F(\cdot,\cdot)$ is a graph alignment distance, as formally shown in Appendix~\ref{app:proof-identified-vertices-01-is-GAS}.} 
$d_F(g_1,g_2):=\norm{\omega(g_1)-\omega(g_2)}{F}$, map $\omega:(\mc G,d_F)\rightarrow([0,1]^{N\times N},\norm{\cdot}{F})$ is an isometry.

Being the co-domain of $\omega$ an Euclidean space, we compute a matrix $X\in \R^{k\times M}$ whose columns $x_i$ constitute a $k$-dimensional vector configuration related to the prototype graphs; this is done via Classical Scaling \cite{pkekalska+duin2005}, that is, $\norm{x_i-x_j}{2}=d(r_i,r_j)$ for all pairs of prototypes $r_i,r_j\in R$. 
As usual, we consider the smallest possible $k$ that preserves the data structure as much as possible.
Successively, for any dissimilarity vector $y\in\mc D$, we define $u:=XJ\,y^{*2}$ to be a linear transformation of $y^{*2}$, obtained by squaring all the components of $y$; the matrix $J$ is the centering matrix $I-\frac{1}{M}\vec 1\vec 1^\top$.
We apply the same procedure of Section~\ref{sec:our-implementation}, considering $u$ instead of $y$: matrix $\Sigma$ is derived from the non-singular covariance matrix\footnote{$\Sigma$ is non-singular, since there are no isolated points in $\mc G$ and as a consequence of the selection of $k$.}, and the statistic $s_w$ is the Mahalanobis distance $d_\Sigma\left(\smean{\vec u}_0,\smean{\vec u}_w\right)$.

Considering the space $(\mc G,d_F)$ and the above transform, we claim that the following lemma holds. Lemma~\ref{lemma:identified-vertices} proves the fulfilment of Assumption (A3).
\begin{lemma}\label{lemma:identified-vertices}
For any positive definite matrix $\Sigma\in\R^{k\times k}$,
$$
c\;d_\Sigma(u_1,u_2) \leq d_F(g_1,g_2)\leq C\;d_\Sigma(u_1,u_2),
$$
where $c=\sqrt{\frac{\lambda_k(\Sigma)}{4\lambda_1(XX^\top)}}$, $C=\sqrt{\frac{\lambda_1(\Sigma)}{4\lambda_k(XX^\top)}}$. $\lambda_i(\cdot)$ is the $i$-th eigenvalue in descending order of magnitude.
\end{lemma}

The proof is reported in Appendix~\ref{sec:proofs-particular}.

\section{Experiments}
\label{sec:experiments}

The proposed methodology can operate on very general families of graphs. Besides the theoretical foundations around which the paper is built, we provide some experimental results showing the effectiveness of what proposed in real change detection problems in streams of graphs.
In particular, we consider the method introduced in Section~\ref{sec:clt-implementation} as an instance of our methodology.
Source code for replicating the experiments is available in \cite{zambon2017cdg}.

\subsection{Experiment description}

\subsubsection{Data}

The experimental evaluation is performed on the well-known IAM benchmarking databases \cite{riesen+bunke2008}.
The IAM datasets contain attributed graphs representing drawings and biochemical compounds. Here we consider the \emph{Letters}, \emph{Mutagenicity}, and \emph{AIDS} datasets. 

The Letters dataset contains 2-dimensional geometric graphs. As such, each vertex of the graphs is characterized by a real vector representing its location on a plane. The edges define lines such that the graphical planar representation of the graphs resembles a latin-script letter. The dataset is composed of 15 classes (one for each letter%
\footnote{The IAM letters database \cite{riesen+bunke2008} considers only non-curved letters; hence, e.g., letters A and E are considered, whereas B and C are excluded.}%
) containing 150 instances each.

Conversely, the Mutagenicity and AIDS datasets contain biological molecules. Molecules are represented as graphs by considering each atom as a vertex and each chemical link as an edge. Each vertex is attributed with a chemical symbol, whereas the edges are labelled according to the valence of the link. Both datasets contain two classes of graphs: mutagenic/non-mutagenic for Mutagenicity and active/inactive for AIDS. The two datasets are imbalanced in terms of size of each classes, in particular Mutagenicity has 2401 mutagenic and 1963 non-mutagenic molecules; AIDS contains 400 active and 1600 inactive molecules.

We considered these datasets because they contain different types of graphs with variegated attributes (numerical and categorical).
We refer the reader to \cite{riesen+bunke2008} and references therein for a more in-depth discussion about the datasets.

\medskip
\subsubsection{Simulating the generating process $\mc P$}
\label{sec:simulating_the_process}

For each experiment in Table~\ref{table:experiments}, we consider two collections of graphs containing all possible observations in the nominal and non-nominal regimes, respectively. 
Each collection is composed by graphs present in one or more predefined classes of the dataset under investigation.
The collections have to be different, but they do not need to be disjoint; as such, some graphs can belong to both collections.

Next, we simulate the process $\mc P$ by bootstrapping graphs from the first collection up to the predefined change time $\tau$; this is the nominal regime. After $\tau$, we bootstrap objects from the second collection hence modelling a change.

Regarding molecular datasets, we considered two distinct experiments. For the Mutagenicity (AIDS) experiment, we set the non-mutagenic (inactive) class as nominal collection and mutagenic (active) class as non-nominal one. On the other side, for the Letter dataset we design four different experiments depending on which classes will populate the collections.
Table~\ref{table:experiments} reports the settings of all the experiments and next section describes the relevant parameters.

\begin{table}
\centering
\footnotesize
\caption{Experimental settings. The first column contains an identifier for each experiment; in particular, in the Letter dataset, `D', `O', and `S' stand for disjoint, overlapping, and subset, respectively. The second column reports the dataset involved, and the third and fourth columns show the set of classes from which nominal/non-nominal graphs are extracted.
The collections of letters are selected in alphabetical order.
%Finally, the last row reports the number $M$ of selected prototypes for the dissimilarity representation.
}
\begin{tabular}{|r|c|cc|}
\hline
 & & \multicolumn{2}{c|}{Collection}\\ 
Experiment ID & Dataset & Nominal & Non-nominal\\ 
\hline
L-D2       &Letter &A,E & F,H  \\
L-D5       &Letter &  A,E,F,H,I    & K,L,M,N,T  \\
L-O        &Letter & A,E,F,H&  F,H,I,K \\
L-S       &Letter & A,E,F,H,I & F,H,I \\
 MUT     &Mutagenicity & non-mutagenic  &mutagenic    \\
 AIDS       &AIDS &   inactive& active  \\
\hline
\end{tabular}
\label{table:experiments}
\end{table}

\medskip
\subsubsection{Parameters setting}
\label{sec:exp:parameters}

For all experiments, the offset $q$ is set to the third quartile of the $\chi^2(M)$ distribution.
The time-dependent threshold $h_w$ has been numerically estimated by Monte Carlo sampling. We drew one million processes of i.i.d.\ random variables $s_w$ by taking the square root of i.i.d.\ random variables distributed as $\chi^2(M)$.
For each obtained process (stream), we computed the sequence of cumulative sums $S_w$ like in \eqref{eq:cusum-iteration} and estimated the threshold $h_w$ as the quantile of order $\alpha_w=$1/ARL0, with ARL0 $=200$ (windows).

We divided the training set $T$ in two disjoint subsets, $T_c,T_p$, used during the prototype selection and change detection learning phases, respectively. We set $|T_p|=300$ and $|T_c|=1000$, afterwards we generated a stream of graphs containing $20\cdot n\cdot$ARL0 observations associated with the operational phase. The change is forced at time $\tau=12\cdot n\cdot\text{ARL0}$. 
As for the distance $d(\cdot,\cdot)$, we considered the bipartite GED implemented in \cite{riesen2013novel}, where we selected the Volgenant and Jonker assignment algorithm. The other GED parameters are set according to the type of graphs under analysis, i.e., for geometric graphs we consider the Euclidean distance between numerical attributes, and a binary (0-1 distance) for categorical attributes. 
The k-Centres procedure is repeated 20 times.

We believe that the selected parameter settings are reasonable. Nevertheless, a proper investigation of their impact w.r.t.\ performance metrics is performed in a companion paper \cite{zambon2017detecting}, hence it is outside the focus of the present one.

\medskip
\subsubsection{Figures of merit}

We assess the performance of the proposed methodology by means of the figures of merit here described. Such measurements are obtained by replicating each experiment one hundred times; we report the average of the observed measures with their estimated 95\% confidence interval (95CI) or standard deviation (std).

First of all, we consider the observed ARL0 introduced in Section~\ref{sec:cd-test} and the delay of detection (DoD). Both of them are computed as the average time lapses between consecutive alarms, but limiting to those alarms raised before and after time $\tau$, respectively.
From them we estimate the rate of detected changes (DCR), by assessing the rate of simulations in which the DoD is less then the observed ARL0.
Finally, we consider also the estimated rate of false anomalies within 1000 samples (FA1000). This is computed as the ratio between the count of raised false alarms and the total number of thousands of time steps under the nominal condition.

We point out that the measures ARL0, DoD, and FA1000 are computed with the window as unitary time step.

\medskip
\subsubsection{Baseline methods}
\label{sec:baseline}

As previously mentioned, state-of-the-art change detection methods for graph streams usually assume a given topology with a fixed number of vertices and/or simple numeric features for vertices/edges.
As reported in \cite{ranshous2015anomaly}, considering a variable topology, a common methodology for anomaly detection on graphs consists of extracting some topological features at each time step and then applying a more general anomaly detector on the resulting numeric sequence. Accordingly, in addition to the method proposed in Section~\ref{sec:our-implementation}, we consider two baseline methods for comparison.
More precisely, we considered two topological features: the density of edges $\phi_1(g)=\frac{|E|}{|V|(|V|-1)}$ and the spectral gap $\phi_2(g)=|\lambda_1(L(g))|-|\lambda_2(L(g))|$ of the Laplacian matrix $L(g)$ \cite{chung1994}.
The particular choices of $\phi_1$ and $\phi_2$ can be justified by considering that both features are suitable for describing graphs with a variable number of vertices and edges.
We implemented two CUSUM-like change detection tests as in Section~\ref{sec:our-implementation} where, for $i=1,2$, the statistic $s_t$ is now given by $\left|\phi_i(g_t)-\expect[\phi_i(g_t)]\right|$, and $\expect[\phi_i(g_t)]$ is numerically estimated in the training phase.

In addition, we consider a further baseline implemented as a degenerate case of our method by selecting only $M=1$ prototype and window size $n=25$. This last baseline is introduced to show that the strength of our methodology resides also in the embedding procedure, and not only in the graph distance $d(\cdot,\cdot)$.

%\medskip
\subsection{Results on IAM graph database}

\newcommand{\smax}{$*\;$}
\newcommand{\strep}{$\circ\;$}
\newcommand{\strew}{$\bullet\;$}
\newcommand{\swin}{$\dagger\;$}

\begin{table*}
\centering
\footnotesize
\caption{Results attained by our method varying number of prototypes ($M$) and window size ($n$). 
Symbol \smax indicate DCR statistically larger than 0.99, with a confidence of 95\%. Symbols \strep and \strew highlight significant trends in the DCR increasing $M$ and $n$, respectively.}
%iam200-12-8
\begin{tabular}{|ccc|rc|cc|cc|cc|}
\hline
\multicolumn{3}{|c|}{Experiment} & \multicolumn{2}{c|}{DCR} & \multicolumn{2}{c|}{ARL0} & \multicolumn{2}{c|}{DoD} & \multicolumn{2}{c|}{FA1000} \\
dataset & $M$ & $n$ & mean & 95CI      & mean & 95CI               & mean & 95CI  & mean & std \\    
\hline
\hline
L-D2 & 4 & 5  &\smax 1.000    & [1.000, 1.000]    & 189   & [103, 333]    & 26    & [8, 66]   & 1.135     & 0.368     \\
\hline
L-D5  & 4 & 5    & 0.380    & [0.290, 0.480]    & 175   & [91, 271]     & 411   & [9, 2149]     & 1.207     & 0.390     \\
L-D5  & 4 & 25   & 0.970   & [0.930, 1.000]    & 183   & [88, 306]     & 41    & [1, 184]  & 0.233     & 0.086     \\
L-D5  & 4 & 125  &\smax 1.000  & [1.000, 1.000]    & 305   & [61, 1148]    & 2     & [1, 5]    & 0.045     & 0.038     \\
L-D5  & 8 & 25   & 0.990   & [0.970, 1.000]    & 175   & [93, 372]     & 10    & [1, 54]   & 0.245     & 0.078     \\
L-D5  & 8 & 125  &\smax 1.000  & [1.000, 1.000]    & 255   & [50, 928]     & 1     & [1, 1]    & 0.054     & 0.047     \\
\hline
L-O   & 4 & 5    & 0.230    & [0.150, 0.310]    & 191   & [120, 344]    & 410   & [21, 1136]    & 1.096     & 0.318     \\
L-O   & 4 & 25   &\strep 0.790   & [0.710, 0.870]    & 205   & [101, 355]    & 123   & [3, 723]  & 0.205     & 0.070     \\
L-O   & 4 & 125  & 0.940  & [0.890, 0.980]    & 291   & [55, 725]     & 37    & [1, 405]  & 0.045     & 0.035     \\
L-O   & 8 & 25   &\strep 0.980   & [0.950, 1.000]    & 163   & [92, 306]     & 24    & [2, 132]  & 0.260     & 0.080     \\
L-O   & 8 & 125  &\smax 1.000  & [1.000, 1.000]    & 238   & [52, 620]     & 2     & [1, 5]    & 0.052     & 0.053     \\
\hline
L-S   & 4 & 5  & 0.720    & [0.630, 0.810]    & 164   & [90, 359]     & 132   & [33, 349]     & 1.321     & 0.418     \\
L-S   & 4 & 25  &\smax 1.000   & [1.000, 1.000]    & 193   & [103, 372]    & 28    & [4, 123]  & 0.223     & 0.075     \\
\hline
AIDS  & 4 & 5  &\smax 1.000    & [1.000, 1.000]    & 175   & [100, 337]    & 1     & [1, 1]    & 1.221     & 0.364     \\
\hline
MUT   & 4 & 5    &\strew 0.050    & [0.010, 0.100]    & 37    & [18, 89]  & 70    & [39, 124]     & 6.294     & 2.159     \\
MUT   & 4 & 25   &\strew 0.800   & [0.720, 0.880]    & 52    & [19, 145]     & 53    & [4, 348]  & 1.075     & 0.521     \\
MUT   & 4 & 125  &\strew 0.980  & [0.950, 1.000]    & 153   & [8, 820]  & 5     & [1, 22]   & 0.276     & 0.284     \\
MUT   & 8 & 25   & 0.920   & [0.860, 0.970]    & 42    & [15, 91]  & 15    & [2, 75]   & 1.242     & 0.680     \\
MUT   & 8 & 125  &\smax 1.000  & [1.000, 1.000]    & 112   & [9, 540]  & 2     & [1, 4]    & 0.292     & 0.267     \\
\hline
\end{tabular}
\label{table:method}
\end{table*}

\begin{table*}
\centering
\footnotesize
\caption{Results attained by baseline methods (index column) based on the graph density (den), the spectral gap of the Laplacian (SG), and the degenerate implementation of the methodology with $M=1$ prototype (M1). Symbol \swin indicates significantly better (95\% confidence) results.}
%iam200-12-8
\begin{tabular}{|ccc|rc|cc|cc|cc|}
\hline
\multicolumn{3}{|c|}{Experiment} & \multicolumn{2}{c|}{DCR} & \multicolumn{2}{c|}{ARL0} & \multicolumn{2}{c|}{DoD} & \multicolumn{2}{c|}{FA1000} \\
dataset & index & $n$ & mean & 95CI      & mean & 95CI               & mean & 95CI  & mean & std \\    
\hline
\hline
L-D2  & M1 &  25  & 1.000   & [1.000, 1.000]    & 238   & [106, 594]    & 15    & [2, 87]   & 0.194     & 0.082     \\
L-D2   & den & 1  & 1.000  & [1.000, 1.000]    & 194   & [68, 556]     & 41    & [22, 104]     & 6.475     & 3.337     \\
L-D2    & SG & 1  & 0.850  & [0.780, 0.920]    & 210   & [75, 394]     & 133   & [55, 274]     & 6.460     & 3.342     \\
\hline
L-D5  & M1 &  25  & 0.780   & [0.700, 0.860]    & 222   & [96, 407]     & 85    & [2, 542]  & 0.200     & 0.098     \\
L-D5   & den & 1  & 0.180  & [0.110, 0.260]    & 219   & [70, 485]     & 369   & [91, 1147]    & 5.775     & 3.346     \\
L-D5    & SG & 1  &\swin 1.000  & [1.000, 1.000]    & 209   & [74, 674]     & 33    & [18, 53]  & 6.058     & 2.971     \\
\hline
L-O   & M1 & 25  & 0.500   & [0.400, 0.600]    & 218   & [100, 459]    & 304   & [7, 1199]     & 0.203     & 0.082     \\
L-O    & den & 1  &\swin 1.000  & [1.000, 1.000]    & 194   & [70, 549]     & 5     & [4, 5]    & 6.853     & 3.428     \\
L-O     & SG & 1  & 0.130  & [0.070, 0.200]    & 221   & [89, 613]     & 352   & [115, 973]    & 5.558     & 2.555     \\
\hline
L-S   & M1 & 25  & 0.880   & [0.810, 0.940]    & 230   & [116, 469]    & 93    & [4, 636]  & 0.189     & 0.066     \\
L-S    & den & 1  &\swin 1.000  & [1.000, 1.000]    & 188   & [79, 385]     & 32    & [20, 48]  & 6.237     & 2.777     \\
L-S     & SG & 1  & 0.080  & [0.030, 0.140]    & 189   & [91, 455]     & 308   & [124, 771]    & 6.100     & 2.247     \\
\hline
AIDS  & M1 & 25  & 1.000   & [1.000, 1.000]    & 229   & [95, 489]     & 1     & [1, 1]    & 0.198     & 0.081     \\
AIDS  & den & 1  & 1.000  & [1.000, 1.000]    & 207   & [67, 505]     & 4     & [2, 6]    & 6.312     & 3.245     \\
AIDS  & SG &  1  & 1.000  & [1.000, 1.000]    & 197   & [79, 443]     & 93    & [48, 152]     & 6.077     & 2.744     \\
\hline
MUT   & M1 & 25  & 0.260   & [0.180, 0.350]    & 154   & [24, 575]     & 484   & [13, 2208]    & 0.506     & 0.441     \\
MUT    & den & 1  & 0.230  & [0.150, 0.310]    & 194   & [69, 381]     & 267   & [96, 741]     & 6.225     & 3.175     \\
MUT     & SG & 1  & 0.030  & [0.000, 0.070]    & 208   & [72, 437]     & 688   & [134, 2342]   & 5.940     & 2.997     \\
\hline
\end{tabular}
\label{table:baseline}
\end{table*}

For the sake of readability, we show results for our method and baselines in two different tables, that is, Table \ref{table:method} and \ref{table:baseline}, respectively.

In all experiments shown in Table~\ref{table:experiments}, there is a parameter setting achieving a detection rate (DCR) statistically larger than $0.99$. Indeed, in Table~\ref{table:method} the 95\% confidence interval (95CI) of the DCR is above $0.99$ (see symbol \smax in the table).
Looking at Table~\ref{table:method} more in detail, we notice that both the window size and the number of prototypes yield higher DCR.
In particular, this can be seen in L-O experiments, where all DCR estimates have disjoint 95CIs (i.e., differences are statistically significant). The same phenomenon appears also for L-D5, L-S, and MUT, e.g., see symbols \strep and \strew in Table~\ref{table:method}.
As far as other figures of merit are concerned, we do not observe statistical evidence of any trend. Still, with the exception of a few cases in MUT, all 95CIs related to ARL0 contain the target value ARL0=200; hence, we may say that the threshold estimation described in Section~\ref{sec:exp:parameters} completed as expected.

Here, we limit the analysis to the proposed parameter settings for $n$ or $M$, since we already reach the highest possible DCR, achieving one hundred detections out of one hundred.
We believe that, by increasing the window size $n$ the false alarms will decrease, as our method relies on the central limit theorem.
Concerning the number $M$ of prototypes, we point out that, in the current implementation of the methodology, the number of parameters to be estimated scales as $M^2$; accordingly, we need to increase the number of samples.

The second experimental analysis addresses the performance assessment of the three baselines of Section~\ref{sec:baseline} on the experiments of Table~\ref{table:experiments}.
The results reported in Table~\ref{table:baseline} show that, in some cases, the considered baselines achieve sound performance, which is comparable to the one shown in Table~\ref{table:method}.
Comparing Table~\ref{table:baseline} with Table~\ref{table:method}, by intersecting the 95\% confidence intervals, we notice that there is always one of the proposed methods which attains DCR that is statistically equivalent or better than the baselines, see symbol \smax in Table~\ref{table:method}.
In particular, the proposed method performs significantly better than the baselines on the MUT dataset.

Finally, Table~\ref{table:baseline} shows that the method based on edge density $\phi_1$ is significantly more accurate in terms of DCR than the one based on spectral gap $\phi_2$ in almost all experiments; confidence intervals at level 95\% do not intersect.
The first method performed better than the degenerate one (M1) with only one prototype: see L-O and L-S. Conversely, M1 outperforms $\phi_1$ on L-D5.

\section{Conclusions}
\label{sec:conclusions}

In this paper, we proposed a methodology for detecting changes in stationarity in streams of graphs. Our approach allows to handle general families of attributed graphs and is not limited to graphs with fixed number of vertices or graphs with (partial) one-to-one correspondence between vertices at different time steps (uniquely identified vertices).
The proposed methodology consists of an embedding step, which maps input graphs onto numeric vectors, bringing the change detection problem back to a more manageable setting.

We designed and tested a specific method as an instance of the proposed methodology.
The embedding has been implemented here as a dissimilarity space representation, hence relying on a suitable set of prototype graphs, which, in our case, provide also a characterization of the nominal condition, and a dissimilarity measure between graphs, here implemented as a graph edit distance.
The method then computes the Mahalanobis' distance between the mean values in two windows of embedded graphs and adopts a CUSUM-like procedure to identify changes in stationarity.

We provided theoretical results proving that, under suitable assumptions, the methodology is able to relate the significance level with which we detect changes in the dissimilarity space with a significance level that changes also occurred in the graph domain; also the vice versa has been proven.
We also showed that our methodology can handle more basic, yet relevant scenarios with uniquely identified vertices.

Finally, we performed experiments on IAM graph datasets showing that the methodology can be applied both to geometric graphs (2-dimensional drawings) and graphs with categorical attributes (molecules), as instances of possible data encountered in real-world applications.
Results show that the proposed method attains at least comparable (and often better) results w.r.t.\ other change detectors for graph streams.

In conclusion, we believe that the proposed methodology opens the way to designing sound change detection methods for sequences of attributed graphs with possibly time-varying topology and non-identified vertices.
In future studies, we plan to work on real-world applications and focus on the automatic optimization of relevant parameters affecting the performance.

%%%%%%%%%%%%%%%%%%%%%%%%%%%%%%%
\appendices

\section{Correspondence between probability spaces}
\label{app:F-is-prob-measure}

Consider the measurable space $(\mc D,\mc B)$ introduced in Section~\ref{sec:methodology:vector}, and the probability space $(\mc G,\mc S,Q)$ of Section~\ref{sec:prob-formulation}.

Let us define the preimage function $\zeta^{-1}(B):=\{g\in\mc G:\zeta(g)\in B\}$, with $\zeta^{-1}(\emptyset)=\emptyset$ and consider the smallest $\sigma$-algebra $\mc S$ containing all open balls%
\footnote{A ball $O(\rho,g)$ is defined as a set $\{f\in\mc G:d(g,f)<\rho\}$ of all graphs $f\in\mc G$ having distance $d(f,g)$ w.r.t.\ a reference graph $g\in\mc G$ smaller than the radius $\rho>0$.}
$O(\rho,g)$ and preimage sets w.r.t.\ any $B\in\mc B$
$$
\{O(\rho,g)|\rho>0, g\in\mc G\}\cup\{\zeta^{-1}(B)|B\in\mc B)\}
$$
and a generic probability density function $Q:\mc S\rightarrow [0,1]$ on $\mc S$.
Then, we can define the function $F:\mc B\rightarrow [0,1]$ as in Equation \eqref{eq:F-from-Q}.

The triple $(\mc D,\mc B,F)$ is a probability space.
The following three properties provide a proof that $F$ is a probability measure on $(\mc D,\mc B)$:
\begin{itemize}
\item $F(\mc D)=Q(\zeta^{-1}(\mc D))=Q(\mc G)=1$, 
\item $F(\emptyset)=Q(\emptyset)=0$, and 
\item for any countable collection $\{B_i\}\in\mc B$ of pairwise disjoint sets, 
$\zeta^{-1}(\cup B_i)=\cup\zeta^{-1}(B_i)$, hence the sets $\zeta^{-1}(B_i)$ are pairwise disjoint and 
$F(\cup B_i)=\cup Q(\zeta^{-1}(B_i))=\cup F(B_i)$.
\end{itemize}

Notice also that, indicating with $\image(\zeta)$ the image set $\{\zeta(g)|g\in\mc G\}\subseteq\mc D$ of $\zeta(\cdot)$, we have $F(B)=0$ for any $B\in\mc B$ such that $B\cap\image(\zeta)=\emptyset$.

\section{Fr\'echet mean}
\label{app:frechet-mean}

Given a probability space $(\mc X,\mc S, P)$ defined on a metric space $(\mc X,d)$, we consider a random sample $\vec x=\{x_{t_1},\dots,x_{t_n}\}$.

\subsection{Definition of Fr\'echet mean and variation}

For any object $x\in\mc X$, let us define the functions $\mc F_{\vec x}(x):=\frac{1}{n}\sum_{t=t_1}^{t_n} d(x,x_t)^2$ and $\mc F_{P}(x):=\int_{\mc{G}} d(x,x')^2\,dP(x')$.
A Fr\'echet sample (population) mean is any object $x$ attaining the minimum of the function $\mc F_{\vec x}(\cdot)$ ($\mc F_{P}(\cdot)$).
We point out that the minimum might not exist in $\mc X$ and, if it does, it can be attained at multiple objects.
Whenever the minimum exists and is unique, we refer to it as $\frmu[\vec x]$ ($\frmu[P]$).
In addition, we define  Fr\'echet sample (population) variation as the infimum $\frvar[\vec x]:=\inf_{x}\mc F_{\vec x} (x)$ ($\frvar[P]:=\inf_{x}\mc F_{P} (x)$).

\subsection{Fr\'echet mean in Euclidean spaces}
\label{app:frechet-classical}

In the case of a set $\mc X\subseteq \R^d$ and distance $d(\cdot,\cdot)=\norm{\cdot-\cdot}{2}$, the space is Euclidean.
Firstly, we show that $\mu[P]=\expect[P]$ and $\mu[\vec x]=\smean{\vec x}$, then we show that 
\begin{equation}\label{eq:expect-variation-euclidean}
\expect[\frvar[\vec x]]=\left(1-\tfrac{1}{n}\right)\,\frvar[P].
\end{equation}

\textbf{1.}
The following equality holds
\begin{equation}\label{eq:frechet-P-fun-rewritten}
\mc F_P(z)=\mc F_P(\expect[P])+\norm{\expect[P]-z}{2}^2.
\end{equation}
as we will show below, in Part \textbf{3}. 
This result proves that the minimum is attained in the expectation $z=\expect[P]$.

\textbf{2.}
Similarly, in the `sample' case
\begin{equation*}%\label{eq:frechet-sample-fun-rewritten}
\mc F_{\vec x}(z)=\mc F_{\vec x}(\smean{\vec x})+\norm{\smean{\vec x}-z}{2}^2,
\end{equation*}
proving that $\mu[\vec x]=\smean{\vec x}$.

\textbf{3.}
The results of previous Parts \textbf{1} and \textbf{2} are derived from the following three equalities: for any $z\in \R^d$
\begin{align*}
    \bullet\, 
        &\norm{a+b}{2}^2 = \norm{a}{2}^2 + 2\,a^\top b + \norm{b}{2}^2,\quad a,b\in \mc X,\\
    \bullet\,
        &\begin{array}{l}
            \int_{\mc X} ({x-\expect[P]})^\top(\expect[P] -z)\;dP(x)=\\
            \quad= \int_{\mc X} x^\top(\expect[P]-z) \;dP(x) - \expect[P]^\top(\expect[P] -z)=0,
        \end{array}\\
    \bullet\,
        &\begin{array}{l}
            \sum\limits_{t} ({x_t-\smean{\vec x}})^\top(\smean{\vec x} -z)= \sum\limits_{t} x_t^\top(\smean{\vec x}-z)-n\,\smean{\vec x}^\top(\smean{\vec x} -z)=0.
            % \\
            % \quad= n\,\smean{\vec x}^\top(\smean{\vec x} -z)- n\,\smean{\vec x}^\top(\smean{\vec x}-z)=0.
        \end{array}
\end{align*}

\textbf{4.} 
Let us move on proving the second result.
Notice that 
\begin{align*}
\frvar[P] &= \expect\left[\norm{x-\frmu[P]}{2}^2\right]\\
&=\expect\left[\norm{x}{2}^2 - 2y^\top\expect[P]+\norm{\frmu[P]}{2}^2\right]\\
&=\expect\left[\norm{x}{2}^2\right] - \norm{\frmu[P]}{2}^2.
\end{align*}

\textbf{5.} 
Then,
\begin{align*}
\frvar[\vec x] &= \tfrac{1}{n}\sum_t \norm{x_t - \frmu[\vec x]}{2}^2
=\tfrac{1}{n}\sum_t x_t^\top x_t - \frmu[\vec x]^\top \frmu[\vec x] \\
& = \tfrac{1}{n}\sum_t x_t^\top x_t - \tfrac{1}{n^2}\sum_{i,j}x_i^\top x_j\\
& = \left(\tfrac{1}{n}-\tfrac{1}{n^2}\right)\sum_t x_t^\top x_t - \tfrac{1}{n^2}\sum_{i\neq j}x_i^\top x_j.
\end{align*}
Now, thanks to the independence of the observations
\begin{multline*}
\expect[\frvar[\vec x]]
=\left(1-\tfrac{1}{n}\right)\,\expect[ x^\top x] - (1-\tfrac{1}{n})\frmu[P]^\top\frmu[P]\\
=\left(1-\tfrac{1}{n}\right)\,\left(\expect[\norm{x}{2}^2] - \norm{\frmu[P]}{2}^2\right)=\left(1-\tfrac{1}{n}\right)\frvar[P],
\end{multline*}
which proves the thesis.

% \end{proof}

\textbf{6.} Finally, in this part, we prove a result that holds in general. 
Notice that $\frac{1}{n}\sum_{t} d(x_t,\frmu[P])^2 \geq \frvar[\vec x]$.
Then, since $\vec x$ are i.i.d.\ so also are the $\{d(x_t,\frmu[P])^2\}$. Thanks to the monotonicity of the expectation, we have 
\begin{equation}\label{eq:expect-variation}
\expect[\frvar[\vec x]]\leq\frac{1}{n}\sum_{t} \expect[d(x_t,\frmu[P])^2]=\frvar[P].
\end{equation}

\section{Proofs of Section~\ref{sec:methodology}}
\label{sec:proofs-general}

\subsection{Proof of Lemma~\ref{lemma:mean-and-zetamean}}
\label{lemma:proof:mean-and-zetamean}

% Let us start with the embedding of the Fr\'echet population mean; the embedding of the Fr\'echet sample mean can be derived in completely similar fashion.

Notice that the means $\frmu[\vec y]=\smean{\vec y}$ and $\frmu[F]=\expect[F]$ are computed w.r.t.\ the Euclidean metric, the and $d'(\cdot,\cdot)$ is deployed as statistic for the change detection test.

\textbf{1.}
From Equation \eqref{eq:frechet-P-fun-rewritten}, for any $z\in\mc D$ we have 
\begin{equation*}%\label{eq:varF-varQ}
\mc F_F(z) = \norm{z- \expect[F]}{2}^2 + \frvar[F].
\end{equation*}

\textbf{2.} 
We provide a second inequality that will be useful later.
Given three graphs $g,f,r\in\mc G$, we have $d(g,f)\geq d(g,r)-d(r,f)$ from the triangular inequality (Assumption (A1)); since this holds for any prototype $r\in R$, it proves that
\begin{equation}\label{eq:norm-equiv-d-inf-2}
d(g,f)\geq \norm{\zeta(g)-\zeta(f)}{\infty}\geq M^{-\frac{1}{2}}\norm{\zeta(g)-\zeta(f)}{2}.
\end{equation}

\textbf{3.} 
Exploiting the inequality \eqref{eq:norm-equiv-d-inf-2} in Part \textbf{2}, and taking $z=\zeta(\frmu[Q])$, we obtain 
\begin{align*}
\frvar[Q]&=\int_{\mc G} d(g,\frmu[Q])^2\,dQ(g)
{=} \int_{\mc G} d(g,\frmu[Q])^2\,dF(\zeta(g))\\
\quad &\geq \int_{\mc G} M^{-1}\,\norm{\zeta(g)- \zeta(\frmu[Q])}{2}^2\,dF(\zeta(g))\\
 &= M^{-1} \int_{\image(\zeta)} \norm{y- \zeta(\frmu[Q])}{2}^2\,dF(y).%=...c^2\mc F_F(\zeta(\frmu[Q]))
\end{align*}
In Appendix~\ref{app:F-is-prob-measure} we observed that $F(\mc D\setminus \image(\zeta))=0$, then
\begin{align*}
\frvar[Q]&\geq 
M^{-1}\int_{\mc D} \norm{y- \zeta(\frmu[Q])}{2}^2\,dF(y)\\ 
&= M^{-1} \mc F_F(\zeta(\frmu[Q])).
\end{align*}
Eventually, combining with Part \textbf{1}, we obtain the first part of the thesis for $v_0:=M\,\frvar[Q]-\frvar[F]\geq 0$, in fact
\begin{equation*}
M\,\frvar[Q]\geq\mc F_F(\zeta(\frmu[Q])) = \norm{\zeta(\frmu[Q])- \expect[F]}{2}^2 + \frvar[F].
\end{equation*}

\textbf{4.} 
Similarly, we have
$$\mc F_{\vec g}(z) = \frvar[\vec y] + \norm{z- \smean{\vec y}}{2}^2$$ 
and, for $v_{\vec g}:=M\,\frvar[\vec g]-\frvar[\vec y]$,
\begin{equation*}
\norm{\zeta(\frmu[\vec g])- \smean{\vec y}}{2}^2 \leq v_{\vec g}.
\end{equation*}

By exploiting \eqref{eq:expect-variation-euclidean} and \eqref{eq:expect-variation}, and the monotonicity of the expected value
$$
\expect[v_{\vec g}]\leq M\,\frvar[Q]-(1-1/n)\frvar[F]=:v_n.
$$

As final remarks, we point out that here we left the dependency from $n$ on purpose, but an independent bound can be easily found, e.g., $v_2=M\,\frvar[Q]-\tfrac{1}{2}\frvar[F]$.

\textbf{5.}
The random variable $\norm{ \smean{\vec y} - \zeta(\frmu[\vec g])}{2}^2$ is non-negative.
As such we can apply Theorem 1, Section 5.4 in \cite{roussas1997course} (sometimes called \emph{Markov's inequality}) and obtain, for each $\delta>0$,
$$
\prob(\norm{ \smean{\vec y} - \zeta(\frmu[\vec g])}{2}^2 \geq \delta )\leq \frac{\expect[\norm{ \smean{\vec y} - \zeta(\frmu[\vec g])}{2}^2]}{\delta}.
$$
Now, from the previous Part \textbf{4} we conclude that
$$
\prob(\norm{ \smean{\vec y} - \zeta(\frmu[\vec g])}{2}^2 \geq \delta )\leq\frac{v_2}{\delta}.
$$

\subsection{Proof of the Lemma~\ref{lemma:meanLS-Phi-ell-u}}
\label{prooflemma:meanLS-Phi-ell-u}

For convenience, let us define the quantities
\begin{equation*}
\begin{array}{rl}
A&   := \{x_0 \in \mc X :d_1(x_0)\leq u(d_2(x_0))\}\\
\rho(-|-)&   := \prob(d_1\leq u(\gamma) \ |\ d_2 \leq\gamma, x\in A )\\
\rho(-|+)&   := \prob(d_1\leq u(\gamma) \ |\ d_2 >   \gamma, x\in A).
\end{array}
\end{equation*}

\textbf{1.} 
By the law of total probability, and for any $\gamma\geq 0$,
\begin{multline*}
\prob(d_1(x)\leq u(\gamma)) =\\
=\prob(d_1(x)\leq u(\gamma)| x \in A) \prob(x \in A) + \\
+\prob(d_1(x)\leq u(\gamma)| x \not\in A) \prob(x \not\in A).
\end{multline*}
Lower-bounding the second addendum with zero and by hypothesis,
$$
\prob(d_1(x)\leq u(\gamma)) 
\ \geq\ \prob(d_1(x)\leq u(\gamma)| x \in A) \cdot p.
$$

\textbf{2.} 
Notice that $\prob(d_1(x)\leq u(\gamma)|d_2(x)=\gamma,x\in A)=1$ for all $\gamma\geq 0$, thanks to the fact that $x\in A$; hence, we have $\rho(-|-)=1$.
Applying again the law of total probabilities,
\begin{multline*}
\prob(d_1(x)\leq u(\gamma)| x \in A) \\
=\rho(-|-) \Phi_2(\gamma)%\prob(d_2(x)< \gamma) 
 +\rho(-|+) (1- \Phi_2(\gamma))%\\%\prob(d_2(x)\geq \gamma) \\
\geq 1\cdot \Phi_2(\gamma).
\end{multline*}
Combining with the above Part \textbf{1}, we have 
$$
\Phi_1(u(\gamma))=\prob(d_1(x)\leq u(\gamma)) \geq p\cdot\Phi_2(\gamma).
$$

\textbf{3.} 
A final remark is that Lemma~\ref{lemma:meanLS-Phi-ell-u} proves also that, if $\ell:\R_+\rightarrow\R_+$ is bijective and increasing providing a lower-bound, then for any $\gamma>0$ and $p> 0$
\begin{multline}\label{eq:remark-on-lemma-with-upperbound-u}
\prob(d_1(x)\geq \ell(d_2(x)))\geq p 
\ \Rightarrow \ 
\Phi_1(\gamma)\leq \frac{1}{p} \cdot \Phi_2(\ell^{-1}(\gamma))
\end{multline}
In fact $d_1\geq \ell(d_2)$ if and only if $d_2\leq \ell^{-1}(d_1)$, therefore, we obtain the result by applying the lemma to $u(\cdot)=\ell^{-1}(\cdot)$ and inverting the roles of $d_1$ and $d_2$.

\subsection{Proof of Proposition~\ref{prop:independent-bounds}}

By Assumption (A2), we can consider the following function of $\vec g$ and their respective CDFs 
$$
\begin{array}{rlc}
d(\vec g)&:=d(\frmu[\vec g],\frmu[Q]),\quad &\Psi(\cdot),\\
d'(\vec g)&:=d'(\zeta(\frmu[\vec g]), \zeta(\frmu[Q])),\quad &\Phi(\cdot),\\
d'_0(\vec g)&:=d'(  \smean{\vec y}, \expect[F]),\quad&\Upsilon(\cdot);
\end{array}
$$
recalling that $\vec y = (\dots,\zeta(g_i),\dots)^\top$.

\textbf{1.} 
From triangular inequality (Assumption (A1)), we have
\begin{multline*}%\label{eq:doppia-triangolare}
     |d'_0(\vec g)-d'(\vec g)|
   \leq 
     d'(\zeta(\frmu[\vec g]), \smean{\vec y}) +  d'( \expect[F], \zeta(\frmu[Q])).
\end{multline*}
By the equivalence of the 1- and 2-norm in $\R^2$,
\begin{multline*}%\label{eq:doppia-triangolare-square}
     (d'_0(\vec g)-d'(\vec g))^2  \leq\\
     \frac{1}{2}\cdot d'(\zeta(\frmu[\vec g]), \smean{\vec y})^2 +   \frac{1}{2}\cdot d'( \expect[F], \zeta(\frmu[Q]))^2.
\end{multline*}
Now, since $d'(\cdot,\cdot)$ is induced by a norm (Assumption (A1)), we can deploy the equivalence of any pair of norms in $\R^M$. Letting $m$ be a constant for relating the distance $d'(\cdot,\cdot)$ with the Euclidean one, and applying Lemma~\ref{lemma:mean-and-zetamean}, we obtain 
\begin{multline*}
\label{eq:abcd1}
(d'_0(\vec g)-d'(\vec g))^2  \leq\\
\leq  \frac{m^2}{2} \left( \norm{\zeta(\frmu[\vec g])- \smean{\vec y}}{2}^2 +  \norm{ \expect[F]-\zeta(\frmu[Q])}{2}^2\right)\\
\leq \frac{m^2}{2}\cdot \left(v_2 + \norm{\zeta(\frmu[\vec g])- \smean{\vec y}}{2}^2\right).
\end{multline*}
By exploiting again Lemma~\ref{lemma:mean-and-zetamean}, for any $\delta>0$, we have 
\begin{multline*}
\prob\left( (d'_0(\vec g)-d'(\vec g) )^2 \leq \frac{m^2}{2}\cdot (v_2 + \delta)\right)\\
\geq \prob\left( \norm{\zeta(\frmu[\vec g])- \smean{\vec y}}{2}^2 \leq \delta\right)\geq  1- \frac{v_2}{\delta}.
\end{multline*}
So, with $b(\delta)=\sqrt{\frac{m^2}{2} (v_2 + \delta)}$ and $p(\delta)=1- \frac{v_2}{\delta}$, we have the following estimates
\begin{equation}
\label{eq:probabilistic-bounds-distance-in-Rn}
\begin{array}{c}
\prob\left(d'_0(\vec g) \leq d'(\vec g)+b(\delta)\right)\geq p(\delta),\\
\prob(d'(\vec g) \leq d'_0(\vec g)+b(\delta))\geq p(\delta).
\end{array}
\end{equation}

\textbf{2.} With Assumption (A3), we can exploit Lemma~\ref{lemma:meanLS-Phi-ell-u} by setting $u(x)=C\,x$, $d_1(\vec g)=d(\vec g)$ and $d_2(\vec g)=d'(\vec g)$. Here $\prob(d_1\leq u(d_2))=1$, hence we obtain that $\Psi(\gamma)\geq \Phi\left(\frac{\gamma}{C}\right)$. Accordingly, employing also \eqref{eq:remark-on-lemma-with-upperbound-u}, we have
$$
\Phi\left(\frac{\gamma}{C}\right)\leq \Psi(\gamma)\leq \Phi\left(\frac{\gamma}{c}\right).
$$

\textbf{3.}
Consider this time \eqref{eq:probabilistic-bounds-distance-in-Rn} with $d_1(\vec g)=d'_0(\vec g)$ and $d_2(\vec g)=d'(\vec g)$.
By applying Lemma~\ref{lemma:meanLS-Phi-ell-u} twice, with $u(x)=x+b(\delta)$ and $\ell(x)=x-b(\delta)$, we obtain
$$
\begin{array}{c}
\Phi(\gamma)\geq  p_\delta\cdot\Upsilon(\gamma-b_\delta)\\
\Phi(\gamma)\leq \tfrac{1}{ p_\delta}\cdot\Upsilon(\gamma+b_\delta)
\end{array}
$$

\textbf{4.} Combining previous parts, Parts \textbf{2} and \textbf{3}, we obtain
$$
\begin{array}{c}
{ p_\delta}
\Upsilon\left(\frac{\gamma}{C}-b_\delta\right)
   \ \leq\ 
        \Psi(\gamma)\ \leq\  
        \frac{1}{ p_\delta} 
        \Upsilon\left(\frac{\gamma}{c}+b_\delta\right).
\end{array}
$$

\section{Proofs of Sections~\ref{sec:our-implementation} and \ref{sec:identified-vertices-01}}
\label{sec:proofs-particular}

\subsection{Proof of Lemma~\ref{lemma:d-mahalanobis-d-graph}}

Recall that $\norm{x}{2}^2\geq\lambda_M x^\top \Sigma^{-1} x$, where $\lambda_M$ is the smallest eigenvalue of the covariance matrix $\Sigma$. 
By the triangular inequality we have \eqref{eq:norm-equiv-d-inf-2}, which leads to $c=\sqrt{\frac{\lambda_M}{M}}$.

\subsection{An instance of graph alignment space}
\label{app:proof-identified-vertices-01-is-GAS}

A graph alignment space is a pair $(\mc G,d)$. According to what defined in Section~\ref{sec:identified-vertices-01}, we explicit an attribute kernel determining a graph alignment distance $d$ acting like $d_F$. 

We consider $\mc A=\mc V\times [0,1]$ as attribute set. Following \cite[Ex.3.4]{jain2016geometry}, for any pair $(x,v)\in \mc A$ we define the feature map $\vec \Phi:(x,v)\mapsto (x,e_v)$, where $e_v\in\R^N$ is zero everywhere except for the position $v$ where it assumes the fixed value $\nu$. The attribute kernel $k:\mc A\times\mc A\rightarrow \R$ is then defined accordingly. We consider a vertex disabled if there are no incoming or outgoing edges, but it has always a label of the form $(0,v)$.

It might happen that the ``cost'' of matching vertices is overcome by the improvement of a better topology alignment. This is something which is not contemplated in the distance $d_F$. For avoiding this, it suffices to set the constant $\nu$ sufficiently large, so that the trivial identity alignment is always the optimal alignment (i.e., vertices are uniquely identified).
Setting $\nu=N^2$ ensures this behaviour and yields an alignment distance $d(\cdot, \cdot)$ acting like $d_F(\cdot, \cdot)$.

\subsection{Proof of Lemma~\ref{lemma:identified-vertices}}
\label{app:identified-vertices}

\textbf{1.}
For every prototye set $R$, the dissimilarity matrix $D(R,R):=[d_F(r_i,r_j)]_{i,j}$ is Euclidean, since it has been generated by using the Euclidean distance.
In general, the dimension of $\R^k$ is $k=N^2$, although it can be lower. We remind that $N$ is the number of vertices assumed for the input graphs.
We use the notation $x_1,\dots,x_M\in\R^k$ to denote the prototypes $r_1,\dots,r_M\in R$ w.r.t.\ the classical scaling process and $X$ is obtained by stacking $[x_1|\dots|x_M]$ as columns (see \cite[Sec. 3.5.1]{pkekalska+duin2005}). 
The minimal number of prototypes for representing $\R^k$ is $M=k+1$, and this holds true only when the matrix $X$ is full rank.
For obtaining the embedding $z$ of a generic graph $g\in\mc G\setminus R$, we compute $y^{*2}:=[d_F(g,r_1)^2,\dots,d_F(g,r_M)^2]^\top$ and we can solve the following linear system with respect to $z$
$$2\, X^\top z = -\left(J\,y^{*2} - J\,D^{*2}\,\vec 1\right),$$
where $J=I - \frac{1}{n}\vec 1\vec 1^\top$ is the centering matrix, $I$ is the identity matrix, and $D^{*2}$ is the component-wise square of $D(R,R)$ (see \cite[Cor. 3.7]{pkekalska+duin2005}).
The solution is unique, provided that the rank of $X$ is $k$.

\textbf{2.}
The differences $\delta z=z_1-z_2$ and $\delta y^{*2}=y^{*2}_1-y^{*2}_2$ are related by the equation $2\,XX^\top\delta z =-XJ \,\delta y^{*2}$. 

Being $\norm{a}{A}^2=a^\top\, A\, a$ we have that 
$4{\norm{\delta z}{(XX^\top)^2}}=4\norm{XX^\top z}{I}={\norm{XJ\,\delta y^{*2}}{I}}$ and
$$
\frac{\norm{\delta z}{I}}{\norm{XJ\,\delta y^{*2}}{\Sigma^{-1}}} = \frac{\norm{\delta z}{I}}{4\,\norm{\delta z}{(XX^\top)^2}} \cdot \frac{\norm{XJ\,\delta y^{*2}}{I}}{\norm{XJ\,\delta y^{*2}}{\Sigma^{-1}}}.
$$ 
We want to show that $c \leq {\norm{\delta z}{I}}/{\norm{XJ\,\delta y^{*2}}{\Sigma^{-1}}} \leq C$, in fact
{$\norm{\delta z}{I}=\norm{\delta z}{2}=d_F(g_1,g_2)$,} and $\norm{XJ\,\delta y^{*2}}{\Sigma^{-1}} = d_\Sigma(XJ\,y_1^{*2},XJ\,y_2^{*2})$.

\textbf{3.}
Applying \cite[Theo.1]{jensen1997bounds}, we can bound ${\norm{a}{A}}/{\norm{a}{B}}$ in terms of the values $\beta$ for which there exists non-null $q \in \R^M$ such that $(A-\beta B)\,q=\vec 0$. 
For the pair $(I,(XX^\top)^2)$ the values $\beta$ are provided by the square of the inverse eigenvalues $\lambda_1(XX^\top),\dots,\lambda_k(XX^\top)$ of $XX^\top$. The eigenvalues are reported in descending order. For what concerns $(I,\Sigma^{-1})$, instead, we see that the values $\beta$ corresponds to the eigenvalues $\lambda_1(\Sigma),\dots,\lambda_k(\Sigma)$ of $\Sigma$. 
In the end, the positive constants $c,C$ are 
$$c^2=\frac{\lambda_k({\Sigma})}{4\,\lambda_1(XX^\top)},\quad C^2=\frac{\lambda_1({\Sigma})}{4\,\lambda_k(XX^\top)}.$$
The non-singularity of $XX^\top$ makes the above fractions feasible.

A final comment on how to exploit Lemma~\ref{lemma:identified-vertices} in Proposition~\ref{prop:independent-bounds} is reported in Appendix~\ref{sec:prop-identified}.

\subsection{Proposition~\ref{prop:independent-bounds} for the special case of Section \ref{sec:identified-vertices-01}}
\label{sec:prop-identified}

When considering identified vertices, while Lemma~\ref{lemma:meanLS-Phi-ell-u} is still valid, Lemma~\ref{lemma:mean-and-zetamean} has to be adapted.
Despite the new embedding $\zeta_0(g) = XJ\zeta(g)^{*2}$ is slightly different from the original one in \eqref{eq:diss_embedding},
$\zeta_0(\cdot)$ and $\mc D_0$ can be treated in a similar way.
In particular, a result equivalent to Lemma~\ref{lemma:mean-and-zetamean} can be proved for $\zeta_0(\cdot)$ by solely adapting Part \textbf{2} of the proof shown in Appendix \ref{lemma:proof:mean-and-zetamean},
\begin{align*}
d_F(g_1,g_2)
=\norm{\delta z}{2}
\geq&\frac{\norm{XX^\top\,\delta z}{2}}{\lambda_1(XX^\top)}
=\frac{\norm{XJ\,\delta y^{*2}}{2}}{\lambda_1(XX^\top)}\\
=& (\lambda_1(XX^\top))^{-1}\norm{u_1-u_2}{2}.
\end{align*}
Paying attention in substituting the map $\zeta(\cdot)$ with $\zeta_0(\cdot)$, the rest of the proof holds with $M=\lambda_1(XX^\top)^2$ thus obtaining the Claims (C1) and (C2), with a final result similar to the one of Proposition \ref{prop:independent-bounds}.

\section*{Acknowledgements}

This research is funded by the Swiss National Science Foundation project 200021\_172671: ``ALPSFORT: A Learning graPh-baSed framework FOr cybeR-physical sysTems''.

%%%%%%%%%%%%%%%
%for references
\bibliographystyle{./IEEEtran}
\bibliography{sample}

%% bibliography

\begin{IEEEbiography}[{\includegraphics[scale=0.15,keepaspectratio]{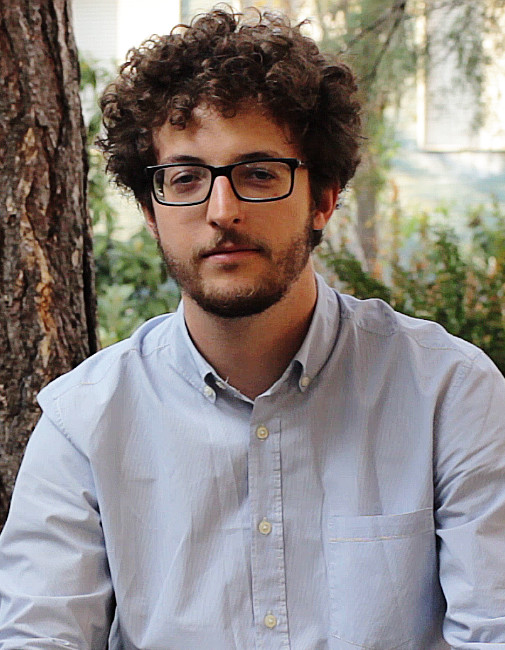}}]{Daniele Zambon}
% Aggiungi qui la tua bibliografia
received the M.Sc. degree in Mathematics from Universit\`a degli Studi di Milano, Italy, in 2016. 
Currently he is Ph.D. student with Faculty of Informatics, Universit\`a della Svizzera italiana, Lugano, Switzerland.
His research interests include: 
graph representation, 
statistical processing of graph streams, 
change and anomaly detection.
\end{IEEEbiography}
\begin{IEEEbiography}[{\includegraphics[scale=0.3,keepaspectratio]{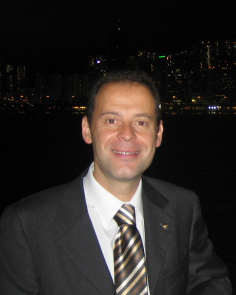}}]{Cesare Alippi}
 (F'06) received the degree in electronic engineering cum laude in 1990 and the PhD in 1995 from Politecnico di Milano, Italy. Currently, he is a Full Professor of information processing systems with Politecnico di Milano, Italy, and of Cyber-Phsical and embedded systems at the Universita' della Svizzera Italiana, Switzerland. He has been a visiting researcher at UCL (UK), MIT (USA), ESPCI (F), CASIA (RC), A*STAR (SIN).
 Alippi is an IEEE Fellow, Distinguished lecturer of the IEEE CIS, Member of the Board of Governors of INNS, Vice-President education of IEEE CIS, Associate editor (AE) of the IEEE Computational Intelligence Magazine, past AE of the IEEE-Trans. Instrumentation and Measurements, IEEE-Trans. Neural Networks, and member and chair of other IEEE committees. 
 In 2004 he received the IEEE Instrumentation and Measurement Society Young Engineer Award; in 2013 he received the IBM Faculty Award. He was also awarded the 2016 IEEE TNNLS outstanding paper award and the 2016 INNS Gabor award.
 Among the others, Alippi was General chair of the International Joint Conference on Neural Networks (IJCNN) in 2012, Program chair in 2014, Co-Chair in 2011. He was General chair of the IEEE Symposium Series on Computational Intelligence 2014. 
 Current research activity addresses adaptation and learning in non-stationary environments and Intelligence for embedded systems. 
 
 Alippi holds 5 patents, has published in 2014 a monograph with Springer on ``Intelligence for embedded systems'' and (co)-authored more than 200 papers in international journals and conference proceedings.
\end{IEEEbiography}
\begin{IEEEbiography}[{\includegraphics[scale=0.4,keepaspectratio]{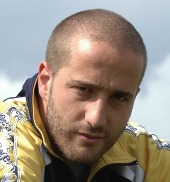}}]{Lorenzo Livi} (M'14) received the B.Sc. degree and M.Sc. degree from the Department of Computer Science, Sapienza University of Rome, Italy, in 2007 and 2010, respectively, and the Ph.D. degree from the Department of Information Engineering, Electronics, and Telecommunications at Sapienza University of Rome, in 2014. He has been with the ICT industry during his studies.
From January 2014 until April 2016, he was a Post Doctoral Fellow at Ryerson University, Toronto, Canada. From May 2016 until September 2016, he was a Post Doctoral Fellow at the Politecnico di Milano, Italy and Universita' della Svizzera Italiana, Lugano, Switzerland.
Currently, he is a Lecturer (Assistant Professor) in Data Analytics with the Department of Computer Science at the University of Exeter, United Kingdom.
He is member of the editorial board of Applied Soft Computing (Elsevier) and a regular reviewer for several international journals, including IEEE Transactions, Information Sciences and Neural Networks (Elsevier). 
His research interests include computational intelligence methods, time-series analysis and complex dynamical systems, with focused applications in systems biology and neuroscience.
\end{IEEEbiography}

\end{document}